\documentclass{article}

\PassOptionsToPackage{numbers, compress}{natbib}

\usepackage[eandd, final]{neurips_2026}

\usepackage[utf8]{inputenc} %
\usepackage[T1]{fontenc}    %
\usepackage{hyperref}       %
\usepackage{url}            %
\usepackage{booktabs}       %
\usepackage{amsmath}
\usepackage{amsthm}
\usepackage{amsfonts}       %
\usepackage{nicefrac}       %
\usepackage{microtype}      %
\usepackage{xcolor}         %
\usepackage{graphicx}
\usepackage{enumitem}
\usepackage{cleveref}
\usepackage{dsfont}
\usepackage{todonotes}
\usepackage{placeins}  %

\theoremstyle{remark}
\newtheorem{remark}{Remark}

\title{CalArena: A Large-Scale Post-Hoc\\ Calibration Benchmark}

\author{%
  Eugène Berta$^{1,2}$\thanks{eugene.berta@inria.fr} \quad
  David Holzmüller$^{1}$ \quad
  Francis Bach$^{1,2}$ \quad
  Michael I. Jordan$^{1,2}$ \\
  $^1$Inria \quad $^2$École Normale Supérieure, PSL Research University
}

\begin{document}

\maketitle

\begin{abstract}
Reliable probability estimates are critical in many machine learning applications, yet modern classifiers are often poorly calibrated. Post-hoc calibration provides a simple and widely used solution, but the large number of proposed methods, combined with small-scale and inconsistent evaluations, makes it difficult to determine which approaches are truly effective in practice.
We introduce a large-scale, standardized benchmark for post-hoc calibration, covering nearly 2000 experiments across tabular and computer vision tasks, including binary, multiclass, and large-scale classification settings. Our benchmark aggregates predictions from a diverse set of classical models, modern deep learning architectures, and foundation models, and provides unified, reproducible implementations of dozens of calibration methods within a common evaluation framework.
We argue that Post-Hoc Improvement (PHI) in proper scoring rules offers a principled alternative to traditional calibration error estimators for comparing post-hoc methods, capturing both calibration quality and potential degradation to the model's predictive performance.
Using this framework, we conduct the most comprehensive empirical study of post-hoc calibration to date.
Our results reveal consistent patterns across domains: smooth calibration functions outperform binning-based approaches, dedicated multiclass methods are essential in high-dimensional settings, and generic machine learning models are not competitive without calibration-specific design. To facilitate future research, we release all data, code, and evaluation tools, providing a plug-and-play benchmark for developing and comparing calibration methods.
\end{abstract}

\begin{figure}[htbp]
    \centering
    \begin{minipage}{\textwidth}
        \centering
        \includegraphics[width=\linewidth]{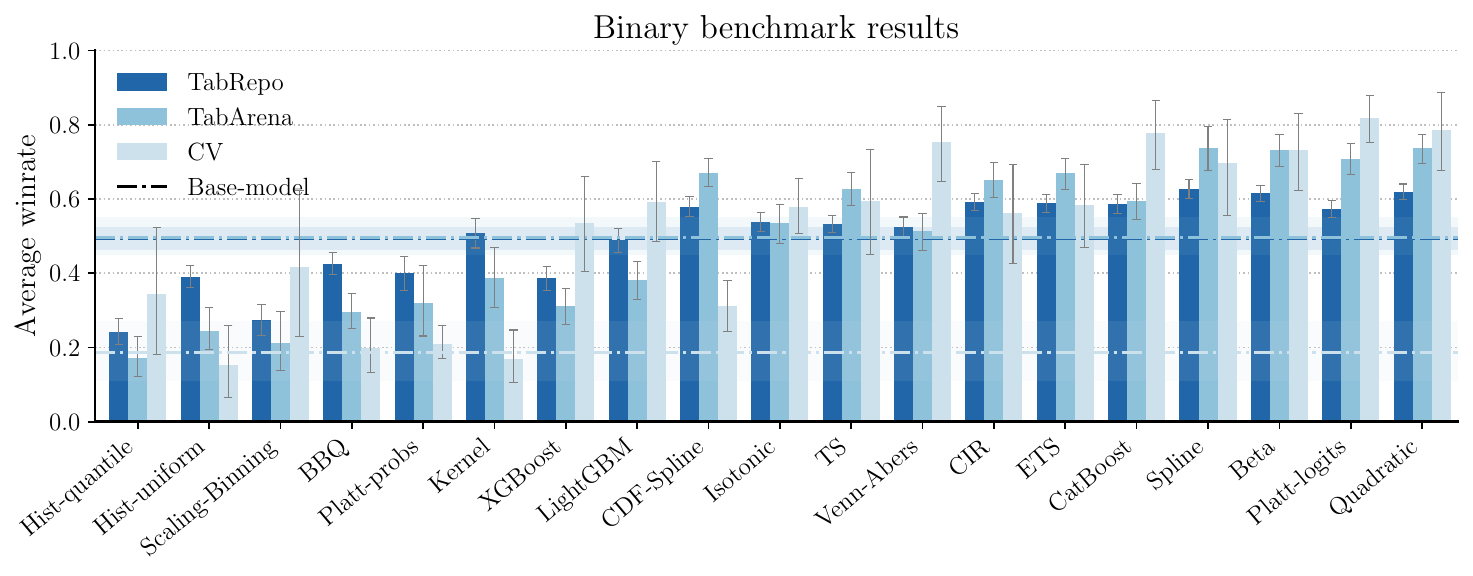}
        \vspace{-3mm}
    \end{minipage}
    \begin{minipage}{\textwidth}
        \centering
        \includegraphics[width=\linewidth]{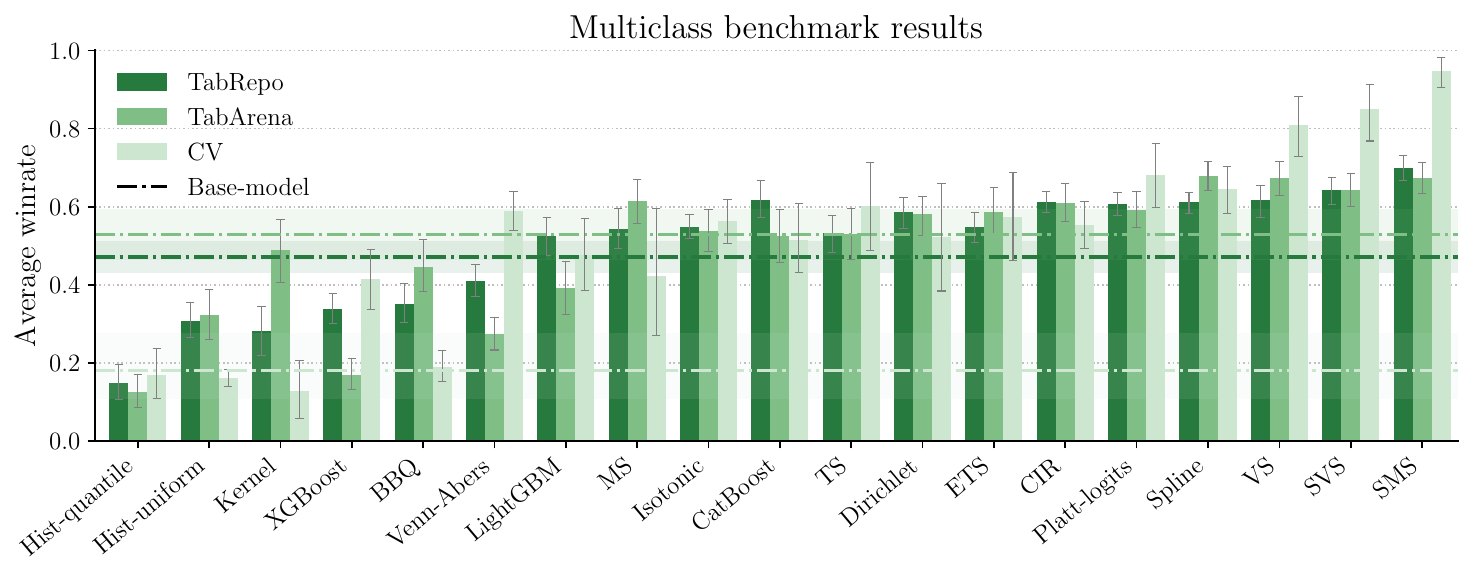}
        \vspace{-3mm}
    \end{minipage}
    \caption{
    Results for binary post-hoc calibration benchmarks \textbf{TabRepo-binary}, \textbf{TabArena-binary} and \textbf{CV-binary} (top) and multiclass post-hoc calibration benchmarks \textbf{TabRepo-multiclass}, \textbf{TabArena-multiclass} and \textbf{CV-multiclass} (bottom).
    Each bar indicates the Brier score winrate, averaged over all experiments in the benchmark with 95\% Confidence Intervals (CIs) constructed by bootstrapping entire datasets (TabArena-binary \& TabRepo) or experiments directly (TabArena-multiclass \& CV).
    Methods are ranked based on the average winrate over the three benchmarks.
    }
    \label{fig:WinratesResults}
    \vspace{-3mm}
\end{figure}

\section{Introduction}

Accurate classification is central to many machine learning applications, ranging from medical diagnosis and fraud detection to autonomous driving and weather forecasting.
Beyond predicting class labels, modern classifiers output probability distributions that reflect their confidence that the instance belongs to each class.
These probabilistic predictions play a critical role in downstream decision-making, especially in high-stakes settings where uncertainty must be explicitly accounted~for.

In practice, however, these predicted probabilities are often poorly calibrated: they do not match observed class frequencies, which undermines the reliability and trustworthiness of machine learning systems.
Miscalibration has been documented across a wide range of models, from classical methods such as support vector machines and boosted trees \citep{platt1999probabilistic, zadrozny2001obtaining, zadrozny2002transforming, niculescu2005predicting} to deep neural networks \citep{guo2017calibration, minderer2021revisiting}.

A widely adopted remedy is \emph{post-hoc calibration} (\Cref{sec:setting}), which adjusts predicted probabilities after training using a ``calibration function'' learned on a held-out validation set.
Over the years, a large number of post-hoc calibration methods have been proposed for both binary and multiclass classification.
Despite this abundance of methods, it remains unclear which approaches are most effective in practice, and under which conditions.

This lack of clarity stems from several key limitations in the current literature.
First, existing empirical evaluations rely on small-scale or outdated benchmarks, limiting their representativeness of modern machine learning settings.
Second, there is no consensus on how to properly evaluate calibration: commonly used metrics such as Expected Calibration Error (ECE) are known to be sensitive to design choices, making comparisons unreliable.
Third, many proposed methods lack accessible, up-to-date implementations, preventing comprehensive and fair comparisons.
As a result, two papers evaluating the same method can reach contradictory conclusions, and practitioners have no reliable basis for choosing a calibration method.

In this work, we address these challenges through the following contributions:
\begin{itemize}[topsep=0mm, itemsep=0mm, leftmargin=5mm]
    \item We introduce a large-scale benchmark for post-hoc calibration, covering a diverse set of predictive settings, including classical and modern models, binary and multiclass tasks, and both tabular and computer vision domains (\Cref{sec:benchmarks}).
    \item We collect, standardize, and evaluate implementations of dozens of post-hoc calibration methods, enabling a comprehensive and reproducible comparison across methods and scenarios (\Cref{sec:methods}).
    \item We propose a well-grounded approach based on proper scores to compare post-hoc calibration methods, evaluating both the reduction in calibration error and degradation to the predictive performance (refinement error) of the initial model (\Cref{sec:metrics}).
    \item We derive actionable insights into the properties of effective calibration methods, providing guidance for practitioners and identifying promising directions for future research (\Cref{sec:takeaways}).
\end{itemize}
In \Cref{fig:WinratesResults}, we report leaderboards obtained by comparing binary and multiclass post-hoc calibration methods on benchmarks targeting different prediction settings (see details in \Cref{sec:results}).

\section{Background and related work}
\label{sec:setting}

\paragraph{Calibration.}
We consider a classifier $f \colon \mathcal{X} \to \Delta_K$ mapping an input $X$ to a vector of predicted probabilities over $K$ classes, where $\Delta_K$ is the probability simplex (for binary classification, we identify $f(X)$ with the probability of the positive class).
The classifier is \emph{calibrated} if its predictions match the conditional frequencies of the labels $\mathbb{P}(Y = k \mid f(X)) = f(X)_k$ for all $k \in \{1, \dots, K\}$.
For instance, among all instances for which a calibrated binary classifier predicts $0.8$, about 80\% belong to the positive class.
This notion of \emph{full} calibration is stronger than \emph{top-class} calibration, which only requires the largest predicted probability to match the accuracy.
Miscalibration is not limited to uniform over- or under-confidence and can take complex, non-monotonic forms.

\paragraph{Post-hoc calibration.}
Post-hoc calibration corrects a trained, frozen classifier $f$ by learning a calibration function $g \colon \Delta_K \to \Delta_K$ on a held-out calibration set $(X_i, Y_i)_{1 \le i \le n_\mathrm{cal}}$, and predicting $g(f(X))$ at test time.
Since $g$ only accesses $f$ through its predictions, post-hoc calibration applies to any black-box model.
Calibration sets are typically small, so $g$ is chosen in a constrained family of functions, either parametric (e.g., Platt or temperature scaling) or nonparametric (e.g., histogram binning or isotonic regression), see \Cref{sec:methods}.
A good calibration function reduces the calibration error of $f$ without degrading its ability to discriminate between classes (its refinement error, see \Cref{sec:metrics}).
We provide an extended background in \Cref{app:background_calibration}.

\paragraph{Empirical studies of calibration.}
\citet{nixon2019measuring} highlight the sensitivity of calibration evaluations to metric design choices, demonstrating that different calibration measures can lead to markedly different conclusions about the effectiveness of recalibration methods.
\citet{tao2024benchmark} performed a large-scale empirical study on calibration properties using Neural Architecture Search, examining, among other things, the reliability of calibration metrics.
These findings highlight the importance of robust evaluation protocols when comparing calibration methods.
\citet{ovadia2019trust} investigate the effect of dataset shift on predictive uncertainty, including calibration.
While not focused on post-hoc calibration, \citet{nado2021uncertainty} provide a standardized benchmarking framework for uncertainty estimation that illustrates the value of shared implementations, evaluation pipelines, and leaderboards.

\paragraph{Benchmarking post-hoc calibration.}
\citet{guo2017calibration} provide one of the earliest widely recognized empirical comparisons of post-hoc calibration techniques on neural network predictions.
\citet{kull2019beyond} released computer vision model predictions that have since been widely reused as a benchmark.
More recently, \citet{berta2026structured} evaluated binary and multiclass calibration methods using tabular predictions from the large-scale repository TabRepo.
In a similar spirit, \citet{manokhin2026classifier} benchmark five calibration techniques on binary tasks from TabArena, a follow-up to TabRepo.

CalArena extends this line of benchmarking efforts by providing a substantially broader experimental scope, covering binary, multiclass, and large-scale multiclass settings across tabular (TabRepo and TabArena) and vision domains, whith a much larger set of calibration methods.
Beyond expanding the scope of experiments, CalArena is designed as a reusable and extensible benchmarking framework, with unified implementations and standardized evaluation protocols intended to facilitate reproducible comparison of both existing and newly proposed post-hoc calibration methods.

\section{Benchmarks}
\label{sec:benchmarks}

We introduce a suite of benchmarks designed to rigorously evaluate post-hoc calibration methods.
These benchmarks cover diverse data modalities (tabular \& computer vision), task types (binary, multiclass, \& large-scale multiclass), and model architectures (classical models, deep learning, \& foundation models).
Benchmarks comprise multiple experiments, each defined as a dataset-model pair with associated validation and test set predictions.
Validation set predictions are used to fit post-hoc calibration methods, while test set predictions allow us to evaluate and compare their performance.

\subsection{Data Sources and Benchmark Construction}

Table~\ref{tab:benchmarks_summary} summarizes the roughly 2000 classification experiments included in our study. To the best of our knowledge, this constitutes the most comprehensive collection dedicated to post-hoc calibration evaluation available in the literature. We construct these benchmarks by consolidating and structuring predictions from several data sources, as detailed below.

\begin{table}[htbp]
    \centering
    \vspace{-1mm}
    \caption{Summary of post-hoc calibration benchmarks constructed.}
    \label{tab:benchmarks_summary}
    \vspace{-1mm}
    \resizebox{\linewidth}{!}{%
    \begin{tabular}{@{}lccccc@{}}
        \toprule
        \textbf{Benchmark Name} & \textbf{Modality} & \textbf{Task} & \textbf{\#Models} & \textbf{\#Datasets} & \textbf{\#Experiments} \\
        \midrule
        TabRepo-binary & Tabular (classical) & Binary & 8 & 104 & 832 \\
        TabArena-binary & Tabular (advanced) & Binary & 11 & 30 & 314 \\
        CV-binary & Vision & Binary & 9 & 3 & 13\textsuperscript{*} \\
        \midrule
        TabRepo-multiclass & Tabular (classical) & Multiclass & 8 & 65 & 520 \\
        TabArena-multiclass & Tabular (advanced) & Multiclass & 11 & 8 & 84 \\
        CV-multiclass & Vision & Multiclass & 10 & 6 & 20 \\
        ImageNet-multiclass & Vision & Large-scale multiclass & 8 & 1 & 8 \\
        \bottomrule
    \end{tabular}%
    }
    \vspace{1mm}
    {\raggedright \footnotesize \textsuperscript{*}Includes eight native binary experiments plus five additional binarized CIFAR-10 experiments.\par}
    \vspace{-3mm}
\end{table}

\paragraph{Classical Tabular Models (TabRepo).} 
We construct the \textbf{TabRepo-binary} and \textbf{TabRepo-multiclass} benchmarks using predictions from \texttt{TabRepo} \citep{salinas2024tabrepo}.
It stores predictions obtained by training classical machine learning and deep learning models on dozens of tabular datasets for a variety of hyperparameter configurations.
We use predictions from eight widely used machine learning models: six classical algorithms (logistic regression, random forests \citep{breiman2001random}, ExtraTrees \citep{geurts2006extremely}, XGBoost \citep{chen2016xgboost}, LightGBM \citep{ke2017lightgbm}, CatBoost \citep{prokhorenkova2018catboost}) and two neural networks from FastAI \citep{howard2020fastai} and AutoGluon \citep{erickson2020autogluon}.
Predictions are stored for 104 binary datasets (yielding 832 experiments) and 65 multiclass datasets (yielding 520 experiments).
We consider one classification task per dataset-model pair and select the hyperparameter configuration that achieves the best validation score (the \emph{tuned} configuration), using the default metrics of \texttt{TabRepo}: ROC-AUC for binary and logloss for multiclass tasks.

\paragraph{Advanced Tabular Models (TabArena).} 
To target the post-hoc calibration of advanced tabular architectures, we create the \textbf{TabArena-binary} and \textbf{TabArena-multiclass} benchmarks using predictions from the large-scale tabular machine learning benchmark \texttt{TabArena} \citep{erickson2025tabarena}.
We select 11 highly competitive models achieving over 1300 Elo on the \texttt{TabArena} leaderboard (as of April 1, 2026, v0.1.3.1), excluding ensembles and models already present in \texttt{TabRepo}.
This selection includes tabular foundation models (RealTabPFN-2.5, TabPFN-2.6 \citep{grinsztajn2025tabpfn}, TabICL, TabICLv2 \citep{qu2025tabicl, qu2026tabiclv2}, LimiX \citep{zhang2025limix}, BetaTabPFN \citep{liu2025tabpfn}, Mitra \citep{zhang2025mitra}, TabDPT \citep{ma2024tabdpt}) and deep learning models (RealMLP \citep{holzmuller2024better}, TabM \citep{gorishniy2025tabm}, ModernNCA \citep{ye2025revisiting}). 
As predictions are not available for every dataset-model pair, selecting the tuned configuration (based on validation ROC-AUC for binary and logloss for multiclass) results in 314 binary experiments across 30 datasets and 84 multiclass experiments across 8 datasets.

\paragraph{Computer Vision (CV) Models.} 
Our vision benchmarks consolidate predictions from deep neural networks provided by \citet{kull2019beyond}, which have become ubiquitous in post-hoc calibration evaluation, and \citet{hekler2025beyond}, which cover more recent computer vision architectures like vision transformers.
All provided logits are converted to probabilities prior to evaluation, so that every benchmark exposes the same interface (see the paragraph below).

The \textbf{CV-binary} benchmark (13 experiments) includes four models trained on the Breast dataset \citep{ALDHABYANI2020104863}, four on the Pneumonia dataset \citep{Kermany2018}, and five experiments generated by binarizing multiclass predictions on CIFAR-10.
We obtain binary predictions by summing the probabilities for all the ``animal'' classes (bird, cat, deer, dog, frog, horse) versus all the ``machine'' classes (airplane, automobile, ship, truck).
This grouping creates a semantically meaningful and approximately balanced binary task.

The \textbf{CV-multiclass} benchmark (20 experiments) spans predictions on CIFAR-10, CIFAR-100 \citep{krizhevsky2009learning}, SVHN \citep{netzer2011reading}, Caltech-UCSD Birds \citep{wah2011birds}, Derma \citep{tschandl2018ham10000}, and OCT \citep{Kermany2018}.

The \textbf{ImageNet-multiclass} benchmark (8 experiments) isolates ImageNet \citep{deng2009imagenet} predictions, targeting the calibration of high-dimensional classification models (1000 classes).

These benchmarks cover a broad spectrum of architectures, including classical convolutional networks (e.g., ResNet \citep{he2016deep}, DenseNet \citep{huang2017densely}, Wide-ResNet \citep{zagoruyko2016wide}, LeNet \citep{lecun2002gradient}, ConvNeXt \citep{liu2022convnet}) and modern vision transformers (e.g., ViT \citep{dosovitskiy2021image}, BEiT \citep{bao2022beit}, Swin \citep{liu2021swin}, EVA \citep{fang2023eva}).

\paragraph{Probabilities rather than logits.}
Every calibration method in our benchmark is fitted on the probabilities predicted by the base model; methods that operate on logits (TS, VS, MS, \dots) are applied to the log-probabilities $\log f(X)$.
This is a deliberate design choice: post-hoc calibration learns a map $g$ on top of a frozen model $f$, and predicted probabilities are the one output that every classifier provides. \texttt{TabRepo} and \texttt{TabArena} only store predicted probabilities, and many models do not expose logits.
Our benchmark thus answers the question: ``given (potentially miscalibrated) predicted probabilities, which post-hoc calibration method should I use?''.
When raw logits are available, calibrating them directly could in principle help logit-based methods, since the softmax discards one dimension of the logit space and saturates very confident predictions in finite precision.
In \Cref{app:logits_ablation}, we evaluate TS, VS and MS on raw logits, on log-probabilities and on log-sum-exp-normalized logits for the CV-multiclass and ImageNet-multiclass benchmarks, for which logits are available.
In these experiments, calibrating log-probabilities is never significantly worse than calibrating logits, and it leaves our conclusions unchanged.

\subsection{Data Availability and Reproducibility}

Our benchmarks are constructed from predictions gathered across large external repositories.
While all original predictions are publicly available, accessing them from \texttt{TabRepo} and \texttt{TabArena} requires downloading hundreds of gigabytes of data, making full reproduction from scratch highly resource-intensive.
To eliminate this barrier, we republish the specific model predictions used in our benchmark on \href{https://huggingface.co/datasets/probkit/CalArena}{Hugging Face}.\footnote{We gratefully acknowledge the original creators for their permission to republish this data. The licenses for the original repositories can be found together with the benchmark files on our Hugging Face dataset.}
Each of our seven benchmarks is a single HDF5 file, with a CSV table listing every experiment it contains (dataset, model, calibration and test set sizes, number of classes, and specific configuration chosen from the original data source) so the scope and composition of each benchmark are fully transparent.
The total download size of our benchmark is 1.71GB.

\paragraph{Evaluating a new calibration method.}
A key design goal of CalArena is to make evaluating a new calibration method as frictionless as possible.
After downloading the benchmark files, a user only needs to implement two methods, \texttt{fit(p\_cal, y\_cal)} and \texttt{predict\_proba(p\_test)}, and register the method in our dedicated \texttt{custom\_calibrators.py} file.
The evaluation script then handles data loading, applies the calibrator to every experiment in the chosen benchmark, computes all metrics, and writes the results to a CSV file.
Running the full benchmark on a single calibrator requires a single command:

{\centering\small
\verb|python run_benchmark.py --benchmark tabrepo-binary --calibrator MyCalibrator|\par}

For large-scale evaluation across all calibrators, we provide SLURM batch scripts that submit one job per calibrator, isolating runtimes and enabling wall-clock comparisons on a compute cluster.

\paragraph{Analysis and visualization utilities.}
Beyond the benchmark runner, we provide a suite of analysis utilities covering the statistical tools used in this paper: bootstrap confidence intervals for winrates, Bradley-Terry Elo ratings \citep{bradley1952rank, chiang2024chatbot}, per-metric absolute improvements over the uncalibrated baseline, and the plotting functions used to generate all figures in this paper.

\paragraph{Call for contributions.}
We designed CalArena specifically to assist researchers in rigorously evaluating new post-hoc calibration techniques.
To foster open community development and well-grounded research, we invite practitioners to contribute their methods directly to our repository or primary calibration package, \href{https://github.com/probkit/probmetrics}{\texttt{probmetrics}}.
By routinely executing the benchmarks on our compute cluster, we intend to maintain a regularly updated leaderboard, establishing a living infrastructure dedicated to the long-term advancement of post-hoc calibration.

The benchmark code is available at \url{https://github.com/probkit/CalArena} and the data can be downloaded from \url{https://huggingface.co/datasets/probkit/CalArena}.

\section{Post-hoc calibration methods}
\label{sec:methods}

In this section we provide a short overview of the post-hoc calibration methods included in our benchmark.
A complete description of each method is provided in \Cref{app:calibration_methods}.
We collect and standardize implementations of every post-hoc calibration method listed below in our open-source calibration package \texttt{probmetrics} at \url{https://github.com/probkit/probmetrics}.

\subsection{Binary methods}

Our benchmark covers a diverse set of binary post-hoc calibration methods, beginning with foundational binning-based techniques.
We include histogram regression using both fixed-width bins (\textbf{Hist-uniform}) and a fixed number of samples per bin (\textbf{Hist-quantile}) \citep{zadrozny2001obtaining}.
We evaluate Bayesian Binning into Quantiles (\textbf{BBQ}) \citep{naeini2015obtaining}, a well-known extension that addresses the sensitivity to the choice of the number of bins by marginalizing over different binning schemes in a Bayesian fashion.
Additionally, we include the hybrid \textbf{Scaling-Binning} approach by \citet{kumar2019verified}, which applies Platt scaling prior to binning.

Next, we evaluate order-preserving methods, exemplified by \textbf{Isotonic} Regression \citep{zadrozny2002transforming}, arguably the most widely adopted nonparametric calibration technique.
To explore improvements over this standard formulation, we also benchmark variants:
Centered Isotonic Regression (\textbf{CIR}) \citep{oron2017centered}, and \textbf{Venn-Abers} calibration \citep{vovk2015large}.

For parametric methods, we feature the widely used Platt Scaling \citep{platt1999probabilistic}, which applies an affine logistic transformation on the initial model's scores.
Following the scikit-learn implementation, we try applying Platt scaling on predicted probabilities directly (\textbf{Platt-probs}), which we compare with the more natural idea of applying the transformation on logits instead (\textbf{Platt-logits}).
We include Temperature Scaling (\textbf{TS}) \citep{guo2017calibration}, as well as recent extensions, including Ensemble Temperature Scaling (\textbf{ETS}) \citep{zhang2020mix} and \textbf{Quadratic} Scaling \citep{berta2026structured}.
Finally, we also consider \textbf{Beta} calibration \citep{kull2017beta}.

Our selection is further diversified by two spline-based methods:
\textbf{Spline} Calibration \citep{lucena2018spline}, which models the recalibration mapping directly, and \textbf{CDF-Spline} \citep{gupta2021calibration}, which maps probabilities by approximating the cumulative distribution function.
We also adapt \textbf{Kernel}-based calibration-error estimation ideas from \citet{popordanoska2022consistent} into a post-hoc Nadaraya-Watson recalibration method.

Finally, we include tree-based post-hoc calibration with \textbf{CatBoost} \citep{prokhorenkova2018catboost}, \textbf{LightGBM} \citep{ke2017lightgbm} and \textbf{XGBoost} \citep{chen2016xgboost} classifiers.

\subsection{Multiclass methods}

For multiclass methods, we begin with natively multiclass calibration methods.
The most widely used option is arguably temperature scaling (\textbf{TS}) \citep{guo2017calibration}.
Vector scaling (\textbf{VS}) and matrix scaling (\textbf{MS}) are extensions with additional parameters that were introduced by \citet{guo2017calibration}.
Ensemble Temperature Scaling (\textbf{ETS}) is another variant that is multiclass compatible \citep{zhang2020mix}.
\citet{kull2019beyond} propose adding regularization to matrix scaling, referring to the resulting method as \textbf{Dirichlet} calibration.
Recently, \citet{berta2026structured} revisited matrix and vector scaling regularization, introducing Structured Matrix Scaling (\textbf{SMS}) and Structured Vector Scaling (\textbf{SVS}).
Finally, the Nadaraya-Watson \textbf{Kernel} estimator has been extended to the multiclass simplex with a Dirichlet kernel \citep{popordanoska2022consistent}.
The tree-based \textbf{LightGBM}, \textbf{XGBoost} and \textbf{CatBoost} calibrators also directly handle multiclass predictions.

Beyond native multiclass methods, a widely used alternative is to apply binary calibration methods in a one-versus-rest (OvR) fashion, meaning that we fit one binary calibration function per class to calibrate the binary probability that the class, rather than any other class, is realized.
Given a new sample, a calibrated probability vector is constructed by evaluating each binary calibrator once and normalizing the vector obtained to sum to one.
We consider several binary methods applied OvR, namely: \textbf{Hist-uniform}, \textbf{Hist-quantile}, \textbf{BBQ}, \textbf{Isotonic}, \textbf{CIR}, \textbf{Venn-Abers}, \textbf{Spline} and \textbf{Platt-logits}.

\section{Metrics}
\label{sec:metrics}

Comparing the calibration performance of these methods on our benchmarks requires addressing the challenge of choosing meaningful metrics for comparison.

\subsection{Calibration metrics}

\paragraph{Calibration error estimators.}
Estimating the true calibration error of a classifier is notoriously hard.
Most commonly, this is quantified by the Expected Calibration Error (ECE), popularized by \citet{naeini2015obtaining} and \citet{guo2017calibration}.
However, subsequent work has demonstrated theoretically and empirically the limits of such binning-based estimators \citep{kumar2019verified, vaicenavicius2019evaluating, roelofs2022mitigating}.
While significant effort has been dedicated to circumventing these limitations with smooth estimators \cite{blasiok2024smooth, popordanoska2024consistent, berta2026variational}, there are often difficulties in extending them to multiclass calibration error and no estimator is widely recognized as satisfactory by the calibration community.
In this section we argue that this issue, while an open challenge in general, does not need to be resolved in the specific setting of comparing post-hoc calibration methods on a fixed benchmark.

\paragraph{Comparing post-hoc calibration methods with proper scoring rules.}
The risk (expected loss) measured with a proper score \citep{gneiting2007strictly} such as the Brier score or logloss, evaluates the general quality of probabilistic forecasts, measuring both calibration error and refinement error \citep{brocker2009reliability}.
Smaller risk can indicate smaller calibration error but also smaller refinement error, making it hard to draw conclusions on whether a model is well calibrated.
In the context of post-hoc calibration (see \Cref{sec:setting}), however, where a function $g \colon \Delta_K \to \Delta_K$ is applied on top of a classifier $f(X) \in \Delta_K$, it is known (see for example Appendix A in \citet{berta2025rethinking}) that the post-hoc transformation cannot decrease the refinement error of the classifier: $\mathrm{Refinement}(g \circ f) \ge \mathrm{Refinement}(f)$.
Specifically, the refinement error of $g \circ f$ is equal to that of $f$ if $g$ is an injection and can be larger otherwise.
A smaller risk after post-hoc calibration thus necessarily comes from a reduced calibration error.
Obviously, the post-hoc transformation might also increase the refinement error, which would also be reflected in the risk of $g \circ f$.
We argue that this should be taken into account when evaluating~$g$.

Considering only calibration error after post-hoc calibration can indeed be misleading.
The easiest way to post-hoc calibrate any classifier $f$ is to make a constant prediction matching the empirical frequency of $Y$ on the calibration set $g(f(X)) = \frac{1}{n_\mathrm{cal}}\sum_{i=1}^{n_\mathrm{cal}} Y_i$.
This produces a roughly calibrated model $g \circ f$ but the discrimination power of the initial model $f$ is completely degraded; the refinement error of $g \circ f$ is maximized.
A post-hoc calibration function~$g$ should be judged by its capacity to reduce the calibration error of the initial classifier, while preserving its refinement error.
This is easily measured by the difference in risk before and after calibration (for any proper loss $\ell$), or Post-Hoc Improvement (PHI, which we denote $\Phi$), which we use as our metric of interest: $\Phi_\ell(g) = \mathbb{E}[\ell(f(X), Y)] - \mathbb{E}[\ell(g \circ f(X), Y)]$.
Empirically, we evaluate $\Phi_\ell$ on the test set $(X_i, Y_i)_{1 \le i \le n_\mathrm{test}}$:

{\centering
$\displaystyle
    \Phi_\ell(g) = \frac{1}{n_\mathrm{test}} \sum_{i=1}^{n_\mathrm{test}} \ell(f(X_i), Y_i) - \frac{1}{n_\mathrm{test}} \sum_{i=1}^{n_\mathrm{test}} \ell(g \circ f(X_i), Y_i) \; .
$\par}

We subtract the risk after post-hoc calibration so that the metric is positively oriented, with larger improvement values indicating better recalibration.
Generally we define $\Phi_s$ for any metric $s$, subtracting the metric evaluated before or after calibration depending on the orientation of $s$ so that $\Phi_s$ is always positively oriented, with positive values indicating improvement (larger accuracy or smaller Brier score for example) after post-hoc calibration and negative values indicating degradation.

For the choice of proper loss $\ell$, we follow \citet{selten1998axiomatic, dimitriadis2024evaluating} and others in the probabilistic forecasting literature in considering that the potentially infinite value taken by logloss is problematic and we favor Brier score, making Post-Hoc Improvement in Brier score ($\Phi_\mathrm{BS}$) the main metric in our benchmark.

\paragraph{Other metrics.}
We also report PHI in logloss ($\Phi_\mathrm{log}$), ECE with 15 bins ($\Phi_\text{ECE-15}$), accuracy ($\Phi_\mathrm{Accuracy}$) and, for binary experiments, the Kuiper calibration metric \citep{imanol2022metrics} ($\Phi_\mathrm{Kuiper}$).
For the multiclass experiments, we report the top-class version of the ECE \citep{kumar2019verified} for which probabilities assigned to the top class only are used for computing the calibration error.
See \Cref{app:absolute_results}.
We stress that $\Phi_\text{ECE-15}$ is reported as a diagnostic, not as a criterion for ranking methods: binned ECE estimators favor binning-based calibration functions and ignore refinement error, so a method can reduce the ECE while degrading the overall quality of the predictions (see \Cref{sec:takeaways}).
Smooth estimators \citep{blasiok2024smooth, popordanoska2022consistent} are more robust but expensive to compute at the scale of our benchmark, and might in turn favor smooth calibration functions; this is why we rank methods with proper scores.

\subsection{Result aggregation}

To aggregate a metric computed for each calibration method on each experiment into a single informative ranking of post-hoc calibration methods, we compute the \emph{winrate} of method $i$ among $m$ methods as the proportion, in $[0,1]$, of competing methods it beats on an experiment.
Denoting $s_j$ the metric of interest for method~$j$, and assuming that larger $s$ is better,

{\centering
$\displaystyle
\mathrm{winrate}(i) = \frac{1}{m-1} \sum_{j=1, j \neq i}^m \mathds{1} (s_i > s_j) \; .
$\par}

We aggregate results by averaging winrates for each method over all experiments in one benchmark.
We compute 95\% confidence intervals (CIs) on the winrate of each calibration method by bootstrapping datasets (when the benchmark contains enough different datasets) or experiments directly.
We present results obtained in \Cref{sec:results}.
A strength of winrates is interpretability: to each calibration method, we assign a score between 0 and 1, estimating the probability that it beats another randomly chosen method from the benchmark on a new experiment.

We complement winrates with three other aggregation methods described in \Cref{app:aggregation}: Elo scores in a Bradley-Terry model (\Cref{app:elo_results}), absolute improvements that account for the scale of $\Phi_\mathrm{BS}$ (\Cref{app:absolute_results}), and a statistical analysis following \citet{demsar2006statistical} (\Cref{app:statistical_analysis}).

\section{Results}
\label{sec:results}

In this section we present average winrates for each post-hoc calibration method on every benchmark where it is applicable.
To put results in perspective, we include predictions from the non-calibrated model and treat them as an independent method (Base-model).
While we only discuss performance here, we report the average runtimes of every calibration method considered in \Cref{app:Runtimes}.

\subsection{Binary benchmarks}

In \Cref{fig:WinratesResults} (top), we plot the leaderboard obtained when aggregating winrates for the \textbf{TabRepo-binary}, \textbf{TabArena-binary} and \textbf{CV-binary} benchmarks, targeting respectively the post-hoc calibration of classical binary classifiers on tabular datasets, advanced binary classifiers on tabular datasets and deep learning binary classifiers on CV datasets.

Performance varies little on the \textbf{TabRepo-binary} benchmark: the maximum winrate is around 0.6 while the base model is around 0.5.
This indicates that no method manages to consistently improve over the non-calibrated model or other post-hoc calibration methods. Platt-probs, XGBoost and binning-based methods even degrade the performance of the initial model.
Three methods slightly outperform the others: Spline calibration, Quadratic scaling and Beta calibration.

On \textbf{TabArena-binary}, results are clearer, with the same three methods achieving more than 0.7 average winrate while the base model is below 0.5. Several methods underperform the base model.

The \textbf{CV-binary} benchmark contains fewer experiments (13) so uncertainty is larger.
Because the number of datasets is small (3), we compute CIs by bootstrapping experiments directly.
One method is above 0.8 winrate while the base model is below 0.2, indicating that post-hoc calibration can yield consistent improvement.
Interestingly, while Quadratic, Spline and Beta are still very good, other methods are also competitive: Platt-logits ranks first, CatBoost third and Venn-Abers (which barely improves upon the base model for tabular benchmarks) fourth.

Averaging winrates across the three benchmarks, Quadratic scaling, Platt scaling on the logits, and Beta calibration emerge as the top-performing methods.
All three apply logistic transformations on log probabilities predicted by the base model, with slightly different parameterizations (see \Cref{app:calibration_methods}).
These results suggest a strong advantage for parametric logistic approaches in binary post-hoc calibration. However, the nonparametric Spline calibration method ranks fourth and performs consistently well across all three benchmarks, indicating that carefully tuned nonparametric approaches may also be capable of achieving state-of-the-art performance in the binary setting.

\subsection{Multiclass benchmarks}

In \Cref{fig:WinratesResults} (bottom), we plot the results obtained for the multiclass calibration benchmarks.
We defer results on our large-scale multiclass benchmark \textbf{ImageNet-multiclass} to \Cref{app:ImageNet_results}.

On the \textbf{TabRepo-multiclass} benchmark, which targets post-hoc calibration of classical tabular models on multiclass classification problems, one method stands out: SMS achieves close to 0.7 winrate while the base model is below 0.5. SVS, which is a less parametrized version of SMS, ranks second.
As in the binary benchmark, binning-based techniques (applied OvR here) are worse than the base model.
This is also the case for XGBoost, Venn-Abers and Kernel.
Not all OvR methods are disappointing, however, as CIR, Spline and Platt-logits perform equivalently to VS and CatBoost, with winrates within 0.01 of each other (ranks three to seven).

On \textbf{TabArena-multiclass}, which mostly contains datasets with few classes, Spline, SMS and VS take the top three places, performing equivalently.
This suggests that in low-dimensional settings good OvR methods can be competitive, and off-diagonal parameters and regularization matter less.

On the \textbf{CV-multiclass} benchmark, the results are more pronounced.
The base model is below 0.2 winrate, showing that post-hoc calibration is effective.
SMS ranks first with close to 0.95 winrate, followed by SVS and VS.
Platt-logits and Spline are the best OvR methods but lag far behind native multiclass methods on this benchmark, which includes high-dimensional datasets such as CIFAR-100.

Averaging winrates across the three benchmarks, SMS emerges as the clear winner, followed by SVS and VS.
Once again, OvR Spline calibration ranks fourth and achieves strong performance across the three benchmarks, although its results are noticeably weaker than the three leading methods on the CV benchmark.
OvR Platt-logits, a simple method with two parameters per class, ranks fifth while being one of the fastest multiclass methods, about two orders of magnitude faster than Spline (see \Cref{app:Runtimes}), which makes it an attractive alternative.
These findings further support the effectiveness of parametric logistic models for recalibration across a broad range of predictive settings.
However, the comparatively poor performance of MS and Dirichlet calibration, despite their $k(k+1)$ parameters, reminds us that increased model flexibility does not necessarily translate into better calibration.
In fact, neither method improves substantially over TS, which uses only a single parameter, highlighting the well-known susceptibility of highly parametrized calibration models to overfitting.

\section{Takeaways}
\label{sec:takeaways}

\paragraph{Smoothness matters.}
Smooth calibration functions outperform binning-based methods on our benchmarks.
This is well illustrated by the performance gap observed between standard isotonic regression and CIR, its piecewise-linear variant.
We also observe this effect on multiclass predictions.
Binning-based methods seem appealing when considering simple calibration error estimators: on the absolute improvement results (\Cref{app:absolute_results}), they rank well for ECE-15.
However, they are detrimental to overall performance, which is revealed by our Brier score benchmark.
We argue that model calibration should not come at the cost of general performance, especially when other techniques effectively reduce calibration error while preserving refinement error, as revealed by our results.

\begin{figure}[htbp]
    \centering
        \begin{minipage}{0.48\textwidth}
        \includegraphics[width=0.9\linewidth]{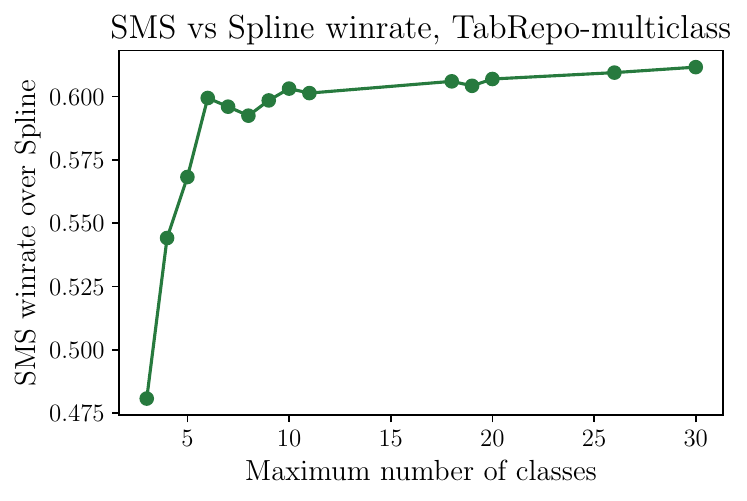}
        \caption{
        Winrate of SMS vs Spline, restricted to datasets with at most~$k$ classes (x-axis).
        }
        \label{fig:SMSvsSpline}
    \end{minipage}
    \hfill
    \begin{minipage}{0.48\textwidth}
        \includegraphics[width=0.9\linewidth]{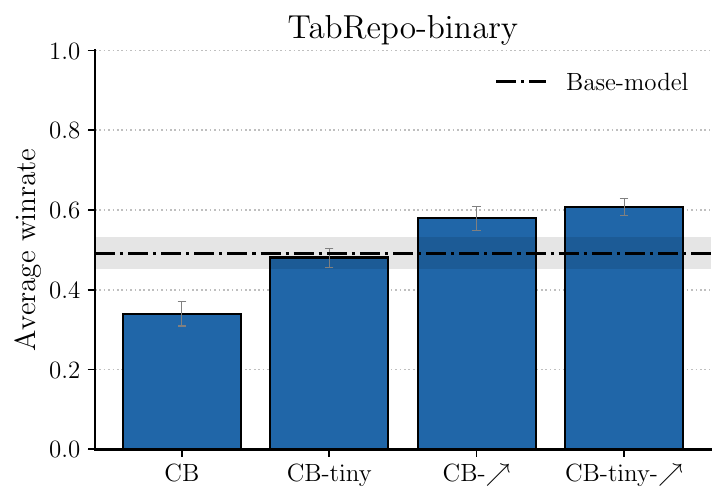}
        \vspace{-2mm}
        \caption{
        Adding calibration design principles to a small CatBoost (CB) classifier improves performance on the \textbf{TabRepo-binary} benchmark.
        }
        \label{fig:CatBoostAblation}
    \end{minipage}

    \vspace{-3mm}
\end{figure}

\paragraph{Native multiclass methods are required for high-dimensional settings.}
While OvR methods demonstrate promising results when the number of classes is small, results on the computer vision datasets, and in particular ImageNet (see \Cref{app:ImageNet_results}), show that native multiclass methods are required to tackle higher-dimensional problems.
To demonstrate this, we plot in \Cref{fig:SMSvsSpline} the winrate of SMS (best multiclass method) against Spline (best OvR method) on the \textbf{TabRepo-multiclass} benchmark when considering datasets with $k$ classes or fewer, where $k$ varies along the x-axis.
As we introduce higher-dimensional datasets in the benchmark, the winrate of SMS increases. It is below 0.5 on datasets with only 3 classes but improves a lot with as few as 4, 5 and 6 classes.

\paragraph{Calibration-specific design is necessary.}
Post-hoc calibration can be framed as a supervised learning problem: predicting a calibrated probability vector from the $K$ uncalibrated probabilities, which suggests using off-the-shelf classifiers such as gradient boosting models.
Yet, even with the (arguably unfair) advantage of early stopping, XGBoost, LightGBM, and CatBoost are generally outperformed by dedicated calibration methods run with default hyperparameters (CatBoost is only competitive on CV-binary and TabRepo-multiclass).
This indicates that generic machine learning models are not well-suited for post-hoc calibration without additional structure, and that calibration-specific design principles are essential.
To illustrate this, we compare on the \textbf{TabRepo-binary} benchmark (\Cref{fig:CatBoostAblation}) a standard CatBoost model (100 trees, default settings) with variants incorporating simple calibration-oriented modifications: reducing the maximum tree depth to three to enforce a lightweight, regularized model and mitigate overfitting (tiny); enforcing monotonicity of the calibrated probabilities with respect to the original predictions, preserving their ranking ($\nearrow$).
Each modification improves performance, and their combination performs best.
This suggests that adapting existing models with calibration-specific constraints is a promising direction for future work.

\paragraph{Flexibility pays off when it is well regularized.}
Given the strong base models in our benchmarks, one might expect only low-variance calibration functions to be competitive.
Our results are more nuanced.
SMS, using the same $K(K+1)$ parameters as matrix scaling, ranks first on average over our multiclass benchmarks, while MS and Dirichlet calibration, without or with less adapted regularization, overfit and barely improve over TS.
Similarly, Spline calibration is among the best methods in both the binary and multiclass settings, performing much better than other nonparametric methods such as histograms or isotonic regression.
Rather than the number of parameters, what distinguishes the best methods is a regularization adapted to the calibration problem: penalties that scale with the number of samples and classes for SMS, and a smoothness penalty for Spline.

\section{Conclusion}
\label{sec:Conclusion}

We presented a large-scale, standardized benchmark for post-hoc calibration, covering nearly 2000 experiments across diverse models, datasets, and prediction settings.
By unifying data sources, implementations, and evaluation protocols, our benchmark provides a reliable and reproducible framework for comparing calibration methods, addressing key limitations of prior empirical studies.

The use of Post-Hoc Improvement in proper scoring rules provides a principled metric for comparing calibration methods, avoiding the pitfalls of traditional calibration error estimators and accounting for both calibration quality and the degradation of predictive performance induced by post-processing.

Our empirical study yields several insights.
Although promising nonparametric alternatives exist, parametric logistic post-hoc calibration models consistently outperform other approaches in both binary and multiclass settings.
Smooth calibration methods also systematically beat binning-based approaches, which often degrade predictive performance despite improving standard calibration error metrics.
One-versus-rest strategies are competitive in low-dimensional settings but fail to scale, making native multiclass methods essential as the number of classes grows.
Finally, generic machine learning models are not competitive out of the box for post-hoc calibration, highlighting the importance of calibration-specific design principles such as adapted regularization and monotonicity.

Beyond these findings, our benchmark is intended as a practical tool for the community to drive ongoing investigation of the relative strengths and weaknesses of various calibration methods.
By releasing all data, code, and evaluation utilities in a plug-and-play framework, we aim to facilitate future research and enable fair, large-scale comparisons of new calibration methods, as well as existing calibration methods that are missing in our benchmark.
We hope this work will contribute to establishing more reliable evaluation standards and accelerate progress in post-hoc calibration.

\paragraph{Limitations.}
Our benchmark evaluates every method with fixed hyperparameters, whereas some could benefit from tuning by cross-validation.
It targets full calibration through proper scores, so rankings could differ for weaker notions such as top-class calibration.
Calibrators are fitted on predicted probabilities, the interface every model provides, rather than on logits; our ablation suggests that this does not affect our conclusions (\Cref{app:logits_ablation}), but this remains to be checked for other methods and data.
Finally, the computer vision benchmarks are smaller than the tabular ones, several recent methods could not be included, and other modalities such as language models remain to be explored.
We discuss these limitations in detail in \Cref{app:Limitations}.

\begin{ack}
We warmly thank Nick Erickson, Markus Kängsepp, Florian Buettner and their co-authors for allowing us to re-publish their model predictions.

This publication is part of the Chair ``Markets and Learning'', supported by Air Liquide, BNP PARIBAS ASSET MANAGEMENT Europe, EDF, Orange and SNCF, sponsors of the Inria Foundation.
This work received support from the French government, managed by the National Research Agency, under the France 2030 program with the reference ``PR[AI]RIE-PSAI'' (ANR-23-IACL-0008).
Funded by the European Union (ERC-2022-SYG-OCEAN-101071601). Views and opinions expressed are however those of the author(s) only and do not necessarily reflect those of the European Union or the European Research Council Executive Agency. Neither the European Union nor the granting authority can be held responsible for them.

\end{ack}

\bibliographystyle{plainnat}
\bibliography{biblio}

\appendix

\section{Background on Probability Calibration}
\label{app:background_calibration}

\subsection{The Concept of Calibration}

In classification tasks, modern machine learning models typically output a probability distribution over the possible classes.
A model is considered \emph{calibrated} if these predicted probabilities accurately reflect the true ground-truth frequencies of the outcomes. 

Formally, for a binary classification problem with input $X \in \mathcal{X}$ and label $Y \in \{0, 1\}$, let $f: \mathcal{X} \to [0, 1]$ be a model predicting the probability of the positive class.
The model is calibrated if the conditional expectation of the target given the prediction equals the prediction itself:
\[
    \mathbb{P}(Y = 1 \mid f(X) = p) = \mathbb{E}[Y \mid f(X) = p] = p, \quad \forall p \in [0, 1] \; .
\]
Empirically, this implies that if we aggregate all instances where a calibrated model predicts a positive-class probability of $0.8$, around $80\%$ of those instances should belong to the positive class. 

Miscalibration occurs when this equality is violated.
Importantly, miscalibration is not limited to systematic, uniform over-confidence or under-confidence across the entire probability space.
It can manifest as complex, non-monotonic patterns where the predicted probability $f(X)$ fails to align with the conditional expectation $\mathbb{E}[Y \mid f(X)]$.
For example, a model might exhibit over-confidence on predictions near the decision boundary (e.g., predicting $0.6$ when the true frequency is $0.5$) while simultaneously being under-confident on extreme predictions (e.g., predicting $0.9$ when the true frequency is $0.99$).

This definition naturally extends to the multiclass setting with $K$ classes, where the model outputs a probability vector $\mathbf{p} = f(X)$ in the probability simplex $\Delta_K$.
Multiclass calibration can be evaluated under varying degrees of strictness.
\emph{Top-class calibration} requires only that the probability assigned to the highest-scoring class matches its empirical accuracy.
In contrast, \emph{full calibration} requires that the entire predicted vector matches the true conditional distribution: $\mathbb{P}(Y = j \mid f(X) = \mathbf{p}) = p_j$ for all $j \in \{1, \dots, K\}$.
Addressing miscalibration is crucial because even highly accurate models often produce unreliable probability estimates, a phenomenon increasingly observed in highly parametrized deep neural networks \citep{guo2017calibration}.

\subsection{Post-Hoc Calibration}

When a base model $f$ exhibits miscalibration, \emph{post-hoc calibration} offers a lightweight, model-agnostic remedy.
Rather than altering the underlying architecture, objective function, or training process of the base model, a secondary calibration function $g: \Delta_k \to \Delta_k$ is learned on an independent, held-out ``calibration set'' $(X_i, Y_i)_{1 \le i \le n_\text{cal}}$.
The objective is to map the uncalibrated outputs to calibrated ones:
\[
    \hat{\mathbf{p}}_{\text{calibrated}} = g(f(X)) \; .
\]
By decoupling the calibration step from the initial model training, post-hoc methods aim to preserve the predictive power, or \emph{refinement error}, of the original classifier while selectively correcting systematic biases in its uncertainty estimates.
As discussed in \Cref{sec:metrics}, if $g$ is an injection, the refinement error of the model remains entirely unchanged.

Because the held-out calibration set is typically much smaller than the primary training dataset, the hypothesis space for $g$ must be constrained to prevent overfitting.
Consequently, post-hoc methods generally rely on low-complexity functions.
These approaches broadly fall into two categories:
\begin{itemize}[topsep=0mm, itemsep=0mm, leftmargin=5mm]
    \item \textbf{Nonparametric methods:} These approaches directly estimate empirical accuracies from partitioned prediction spaces or empirical cumulative distribution functions, typically subject to monotonicity constraints to preserve the ranking of the initial predictions. Prominent examples include Histogram Binning \citep{zadrozny2001obtaining} and Isotonic Regression \citep{zadrozny2002transforming}.
    \item \textbf{Parametric scaling methods:} These methods apply learnable, continuous transformations directly to the model's logits or probabilities. They operate under specific distributional assumptions and include foundational techniques like Platt Scaling \citep{platt1999probabilistic} for binary classification and Temperature Scaling \citep{guo2017calibration} for multiclass classification.
\end{itemize}

\section{Post-hoc calibration methods}
\label{app:calibration_methods}

\subsection{Binary methods}

We denote by $p$ the probability assigned by the initial model to the positive class $Y = 1$.
We denote by $\sigma \colon x \mapsto (1+e^{-x})^{-1}$ the sigmoid function and $\sigma^{-1} \colon x \mapsto \log(x/(1-x))$ its inverse.

\paragraph{Temperature scaling} \citep{guo2017calibration} is probably the most widely used post-hoc calibration method. It uses a single ``temperature'' parameter $T$ to re-scale the logits of the initial classifier before feeding them again through a sigmoid function. The mapping learned in the binary case is
\[
\mathrm{TS}(p) = \sigma( \sigma^{-1}(p) / T ) \; .
\]
The temperature parameter $T$ is chosen to minimize the logloss on the calibration set. As discussed recently by \citet{berta2025rethinking}, this is equivalent to the ``linear scaling'' model
\[
\mathrm{TS}(p) = \sigma( \alpha \sigma^{-1}(p) ) \; ,
\] with the advantage that the logloss minimization problem is convex in $\alpha$. The implementation proposed in the \texttt{probmetrics} package uses this formulation and finds the optimal $\alpha$ via a bisection search on the gradient of the loss on the calibration set. We use this implementation in our benchmark.

\paragraph{Ensemble temperature scaling} \citep{zhang2020mix} (ETS) extends standard temperature scaling by combining multiple simple calibration models into a convex ensemble.
Specifically, ETS forms a weighted average of three components: the original (uncalibrated) probability $p$, the temperature-scaled probability $\mathrm{TS}(p)$, and a uniform distribution baseline,
\[
 \mathrm{ETS}(p) = w_1 \mathrm{TS}(p) + w_2 p + w_3 \frac{1}{2} \; ,
\]
where the weights $w_1, w_2, w_3 \geq 0$ satisfy $w_1 \! + \! w_2 \! + \! w_3 \! = \! 1$. The temperature parameter (for $\mathrm{TS}$) and the mixture weights are jointly optimized on the calibration set by minimizing the logloss. This formulation can be interpreted as a regularized extension of temperature scaling, where the additional components help mitigate overfitting and improve robustness, particularly when the calibration set is small. We adapt the implementation provided in the original paper in the \texttt{probmetrics} package.

\paragraph{Platt scaling} \citep{platt1999probabilistic} fits a sigmoid (affine logistic) model on model predictions.
While it was initially defined for SVM scores, it is commonly used in the literature for other classifiers.
For a probabilistic classifier making predictions in the $[0,1]$ interval, Platt scaling can be applied on the positive class probabilities
\[
\textbf{Platt-probs}(p) = \sigma(\alpha p + \beta)
\]
where the parameters $\alpha$ and $\beta$ are chosen to minimize the logloss on the calibration set. We call this \textbf{Platt-probs} in our benchmark, and use the implementation provided in the \texttt{scikit-learn} package.

As for temperature scaling, it is more natural to apply the logistic model on the logits $\sigma^{-1}(p)$, rather than on probabilities,
\[
\textbf{Platt-logits}(p) = \sigma(\alpha \sigma^{-1}(p) + \beta) \; .
\]
We use the implementation in the \texttt{probmetrics} package.

\paragraph{Quadratic scaling} \citep{berta2026structured} also fits a logistic model but uses a quadratic function of the logits
\[
\mathrm{QS}(p) = \sigma( \gamma \sigma^{-1}(p)^2 + \alpha \sigma^{-1}(p) + \beta) \; .
\]
Platt scaling is originally obtained by modeling the score distributions of the initial model $f$ as normal distributions with equal variance.
The authors observe that modeling the additional flexibility of normal score distributions with different variances requires a quadratic function of the logits, leading to the proposed three-parameter model.
We use the implementation in the \texttt{probmetrics} package.

\paragraph{Beta calibration} \citep{kull2017beta} models the probabilities predicted by the initial model for each class by a Beta distribution on the $[0,1]$ interval, further requiring that the resulting calibration map is non-decreasing. This can also be framed as a logistic model, written as
\[
\mathrm{Beta}(p) = \sigma( a \log(p) - b \log(1-p) + c ) \; ,
\] with free parameters $a,b$ and $c$.
We use the implementation from the \texttt{betacal} package provided by the original paper.

\begin{remark}
    Notice that if we constrain $a = b$, we get $\mathrm{Beta}(p) = \sigma( a \log(p) - a \log(1-p) + c ) = \sigma( a \log(p / (1-p)) + c ) = \sigma( a \sigma^{-1}(p) + c )$ and we recover Platt scaling. This simpler model is called $\mathrm{Beta}[a=b]$ by the authors.
\end{remark}

\paragraph{Histogram regression} \citep{zadrozny2001obtaining} is a nonparametric method that partitions the unit interval $[0, 1]$ into $M$ mutually exclusive bins $B_1, \dots, B_M$.
The calibrated probability assigned to an input is the empirical estimate of $\mathbb{P}(Y=1 \mid p \in B_m)$, which is the fraction of positive samples from the calibration set that fall into the corresponding bin.
Formally, if we denote by $I_m$ the indices of the calibration samples whose initial probability $p_i$ falls into bin $B_m$, the mapping is a piecewise constant step function:
\[
\mathrm{Hist}(p) = \sum_{m=1}^M \theta_m \mathbb{I}(p \in B_m) \; ,
\]
where $\mathbb{I}$ is the indicator function and $\theta_m$ is the empirical accuracy within bin $B_m$.
If a bin contains samples ($|I_m| > 0$), $\theta_m = \frac{1}{|I_m|} \sum_{i \in I_m} Y_i$.
To ensure the function remains defined for sparsely populated regions, if a bin is empty ($|I_m| = 0$), $\theta_m$ defaults to the global prior probability of the positive class over the entire calibration set.
We consider two standard variants for defining the bin boundaries:
\begin{itemize}[topsep=0mm, itemsep=0mm, leftmargin=5mm]
    \item \textbf{Uniform bins:} The probability space is divided into $M$ uniformly sized sub-intervals (e.g., $[0, 1/M), [1/M, 2/M), \dots, [(M-1)/M, 1]$), with the final bin closed on the right to include exact $1.0$ predictions.
    This strategy can yield highly fluctuating estimates if the initial model's probabilities are skewed, leading to sparsely populated bins.
    \item \textbf{Quantile bins:} The boundaries are chosen based on the $M$-quantiles of the initial probabilities $p$ on the calibration set.
    This equal-mass approach aims to ensure that every bin contains approximately the same number of samples, thereby bounding the variance of the empirical estimates $\theta_m$.
    If the initial probabilities contain significant point masses (e.g., identical predictions from tree-based models), duplicate quantiles are merged, resulting in $M' \le M$ empirical bins.
\end{itemize}
We provide a new implementation in the \texttt{probmetrics} package, which supports both binning strategies, and set the initial target number of bins to ten for both methods.

\paragraph{Bayesian binning into quantiles} \citep{naeini2015obtaining} (BBQ) is an ensemble method designed to overcome the high sensitivity of standard histogram regression to the arbitrary choice of bin boundaries and the total number of bins.
Instead of relying on a single partitioning scheme, BBQ considers a vast space of possible equal-mass binning models.
For a given initial probability $p$, each candidate binning model $M$ provides an empirical probability estimate $\mathrm{Hist}_M(p)$. BBQ computes the final calibrated probability as the weighted average of the estimates from all considered models:
\[
\mathrm{BBQ}(p) = \sum_{M} P(M|\mathcal{D}) \mathrm{Hist}_M(p) \; ,
\]
where the weights $P(M|\mathcal{D})$ are the posterior probabilities of each binning scheme given the calibration data $\mathcal{D}$, computed using the marginal likelihood.
We use the implementation provided in the \texttt{netcal} package.

\paragraph{Isotonic regression} \citep{zadrozny2002transforming} is a nonparametric calibration method that learns a piecewise constant, non-decreasing mapping from the initial probabilities to the calibrated ones. Unlike parametric methods that impose a strict functional form, it minimizes the mean squared error (Brier score) between the transformed probabilities and the true labels $Y \in \{0, 1\}$ on the calibration set, subject only to a monotonicity constraint. Formally, it finds a non-decreasing step function $m$ that minimizes
\[
\sum_{i=1}^N (Y_i - m(p_i))^2 \; ,
\]
where $N$ is the number of calibration samples. This optimization problem is solved efficiently using the Pool-Adjacent-Violators Algorithm (PAVA) \citep{ayer1955empirical}. The learned mapping ensures that the rank order of the initial predictions is preserved: if $p_i \leq p_j$, then $m(p_i) \leq m(p_j)$.
We use the implementation provided in the \texttt{scikit-learn} package.

\paragraph{Centered isotonic regression} \citep{oron2017centered} (CIR) extends standard isotonic regression.
When resolving monotonicity violations, PAVA groups adjacent samples into blocks, producing a step function with flat, piecewise-constant intervals.
CIR assumes that this plateauing behavior is undesirable for probability calibration, as it maps distinct initial predictions to the same calibrated score, discarding the strict rank order of the initial model.
CIR modifies this approach by assigning the empirical accuracy $\bar{Y}_k$ of a given block $k$ strictly to the center (the empirical mean) of its initial probabilities, denoted $\bar{p}_k$.
For any initial probability $p$ falling between two adjacent block centers $\bar{p}_k$ and $\bar{p}_{k+1}$, the calibrated probability is obtained via linear interpolation:
\[
\mathrm{CIR}(p) = \bar{Y}_k + \frac{p - \bar{p}_k}{\bar{p}_{k+1} - \bar{p}_k} (\bar{Y}_{k+1} - \bar{Y}_k) \; .
\]
This adjustment yields a strictly increasing, continuous piecewise-linear calibration mapping, avoiding the flat regions of standard isotonic regression everywhere except possibly at the boundaries.
We use the implementation provided by the \texttt{cir-model} Python package.

\paragraph{Venn-Abers predictors} \citep{vovk2015large} provide a theoretical framework for calibration that yields valid probability intervals. For a new test sample with an initial prediction $p$, the method augments the calibration set with this new sample twice: first by assigning it a pseudo-label $Y=0$, and then by assigning it $Y=1$. It fits an isotonic regression model on both augmented sets, yielding two calibrated probabilities, $p_0$ and $p_1$. These two values form a multi-probabilistic prediction interval $[p_0, p_1]$. To obtain a single point estimate for standard evaluation in our benchmark, we combine them into a single probability:
\[
\mathrm{VA}(p) = \frac{p_1}{1 - p_0 + p_1} \; ,
\]
following the canonical probabilistic merging proposed in the original work.
We use the implementation provided in the \texttt{venn-abers} Python package.

\paragraph{Spline calibration} \citep{lucena2018spline} is a nonparametric method that directly models the calibration mapping using smoothing splines. Rather than imposing a rigid parametric form like a sigmoid function, it finds a smooth function $S$ that maps the initial probabilities $p$ to the calibrated ones, where $S$ is typically a cubic smoothing spline. The spline is fitted to minimize a regularized objective function (such as the logloss or Brier score on the calibration set) augmented with a roughness penalty $\lambda \int [S''(x)]^2 dx$. This penalty restricts the curvature of the mapping, preventing the model from overfitting to local noise. This provides a flexible middle ground between the rigid assumptions of parametric scaling and the flat, piecewise-constant plateaus of histogram binning.
We use the implementation provided in the \texttt{splinecalib} Python package.

\paragraph{CDF Spline calibration} \citep{gupta2021calibration} takes a different approach by shifting the spline fitting process from the direct probability space to the cumulative distribution space. The authors observe that while the direct empirical mapping from initial scores to accuracy is inherently noisy and difficult to regularize, its cumulative sum is strictly increasing and much smoother. For a given quantile $t \in [0, 1]$ of the initial probabilities on the calibration set, let $s(t)$ be the corresponding score. The authors define the cumulative function $h(t) = \mathbb{P}(Y = 1, p \leq s(t))$. Because $\mathbb{P}(p \leq s(t)) = t$ by definition of a quantile, it can be shown that the true conditional probability of the positive class given a specific score $s(t)$ is exactly the derivative of $h$. The method fits a cubic spline to the empirical points of $h(t)$ using simple least-squares, and the calibrated probability is analytically recovered by taking its first derivative:
\[
\text{CDF-Spline}(s(t)) = h'(t) \; .
\]
The core difference between the two methods lies in their optimization targets. While standard spline calibration relies on explicit roughness penalties to fit the target probabilities directly, CDF Spline achieves a smooth calibration curve by exploiting the natural stability of the cumulative distribution function and obtaining the final mapping via differentiation.
We use the implementation provided by the authors in the \texttt{spline-calibration} Python package.

\paragraph{Scaling-binning} \citep{kumar2019verified} is a hybrid approach that combines the variance reduction of parametric scaling with the bias reduction of nonparametric binning.
The authors observe that parametric methods (like Platt scaling) are highly sample-efficient but can introduce systematic bias if the true calibration mapping does not perfectly follow a logistic curve. 
Conversely, nonparametric methods (like histogram regression) are expressive but suffer from high variance when data is sparse.
Scaling-binning operates in two sequential steps. First, a parametric scaling function (typically Platt scaling) is fitted to the initial probabilities to produce intermediate scores $p' = \mathrm{PS}(p)$.
Second, a histogram binning scheme is applied to these intermediate scores to correct any residual calibration errors. The final mapping is the composition of the two:
\[
\textrm{Scaling-Binning}(p) = \mathrm{Hist}(\mathrm{PS}(p)) \; .
\]
By using a parametric method as an inductive bias, the subsequent binning step requires fewer bins to achieve optimal calibration.
The authors mathematically demonstrate that this composition yields tighter finite-sample theoretical guarantees on the true calibration error compared to using either method independently.
In our benchmark, we implement this by sequentially chaining the respective scaling and binning modules.
We use the implementation provided in the \texttt{uncertainty-calibration} Python package.

\paragraph{Kernel-based calibration} with a Beta Kernel \citep{popordanoska2022consistent} relies on a nonparametric Nadaraya-Watson estimator to map uncalibrated probabilities to calibrated ones. Because binary probabilities are bounded, standard density estimators like Gaussian kernels typically suffer from boundary bias. To circumvent this, a Beta kernel is employed.
The calibrated probability for a new prediction $p$ is computed as:
\[
\mathrm{Kernel}(p) = \frac{\sum_{i=1}^N Y_i K_h(p, p_i)}{\sum_{i=1}^N K_h(p, p_i)} \; ,
\] where $K_h(p, p_i)$ is the Beta probability density function evaluated at $p$ with shape parameters $\alpha = p_i / h + 1$ and $\beta = (1 - p_i)/h +1$.
The bandwidth parameter $h>0$ controls the smoothness of the estimator.
For computational efficiency, our implementation randomly subsamples the calibration set to a maximum of 10,000 points and determines the bandwidth via an adapted Scott's rule heuristic for one-dimensional inputs, $h=N^{-2/5}$.
We release our implementation in the \texttt{probmetrics} package.

\paragraph{Tree-based calibration.} Post-hoc calibration can be framed as a supervised learning problem: given a $K$-dimensional input (the uncalibrated probabilities), predict a calibrated $K$-dimensional probability vector.
This perspective suggests using off-the-shelf classifiers such as gradient boosting models.
In our benchmark, we evaluate \textbf{LightGBM} \citep{ke2017lightgbm}, \textbf{XGBoost} \citep{chen2016xgboost}, and \textbf{CatBoost} \citep{prokhorenkova2018catboost} as post-hoc calibration functions.
Given that the amount of calibration data is usually small, it is expected that out-of-the-box classifiers would overfit the calibration set.
To mitigate this, we restrict the maximum tree depth to 3 while keeping all other parameters at their default values.
For each method, we train an ensemble of five classifiers via 5-fold cross-validation, using out-of-fold data for early stopping to further prevent overfitting.
Notice that this is an arguably unfair advantage over other methods in our benchmark, which are applied with default parameters and could benefit from parameter tuning that is made possible with cross-validation.
We release these three calibrators in the \texttt{probmetrics} package and refer to the implementations for additional details.

\subsection{Multiclass methods}
\label{app:multiclass_calibration_methods}

We consider a $K$-class classification problem. Let $\mathbf{p} = (p_1, \dots, p_K) \in \Delta_K$ denote the vector of predicted class probabilities from the initial model, where $\Delta_K$ is the probability simplex, such that $\sum_{k=1}^K p_k = 1$. Let $\mathbf{z} = (z_1, \dots, z_K) \in \mathbb{R}^K$ denote the corresponding log probabilities, obtained via $z_k = \log(p_k)$ such that $p_k = \mathrm{softmax}(\mathbf{z})_k = \exp(z_k) / \sum_{j=1}^K \exp(z_j)$.

\paragraph{Temperature scaling} \citep{guo2017calibration} (TS) is the most widely used multiclass post-hoc calibration method.
It rescales the logits of the model by a single scalar temperature parameter $T > 0$, shared across all classes.
The calibrated probabilities are obtained by applying the softmax function to the rescaled logits:
\[
\mathrm{TS}(\mathbf{p})_k = \frac{\exp(z_k / T)}{\sum_{i=1}^K \exp(z_i / T)} \; .
\]
The temperature parameter is learned by minimizing the logloss on the calibration set. This method preserves the ranking of the logits and thus does not change the predicted class.
We use the implementation from the \texttt{probmetrics} package, which learns a scaling parameter $\alpha \times z_k$ instead, making the problem convex \citep{berta2025rethinking}.
The optimal scaling is found by bisection search on the gradient of the loss on the calibration set.

\paragraph{Ensemble temperature scaling} \citep{zhang2020mix} (ETS) extends standard temperature scaling by combining multiple simple calibration models into a convex ensemble.
Specifically, ETS forms a weighted average of three components: the original (uncalibrated) probability $\mathbf{p}$, the temperature-scaled probability $\mathrm{TS}(\mathbf{p})$, and a uniform distribution baseline,
\[
 \mathrm{ETS}(\mathbf{p}) = w_1 \mathrm{TS}(\mathbf{p}) + w_2 \mathbf{p} + w_3 \frac{1}{K} \; ,
\]
where the weights $w_1, w_2, w_3 \geq 0$ satisfy $w_1 \! + \! w_2 \! + \! w_3 \! = \! 1$.
The temperature parameter (for $\mathrm{TS}$) and the mixture weights are jointly optimized on the calibration set by minimizing the logloss.
This formulation can be interpreted as a regularized extension of temperature scaling, where the additional components help mitigate overfitting and improve robustness, particularly when the calibration set is small.
We adapt the implementation provided in the original paper in the \texttt{probmetrics} package.

\paragraph{Vector scaling} \citep{guo2017calibration} (VS) generalizes temperature scaling by introducing a class-specific scaling parameter for each logit. The calibrated probabilities are given by
\[
\mathrm{VS}(\mathbf{p})_k = \frac{\exp(a_k z_k + b_k)}{\sum_{i=1}^K \exp(a_i z_i + b_i)} \; ,
\]
where $\mathbf{a} \in \mathbb{R}^K$ and $\mathbf{b} \in \mathbb{R}^K$ are learned parameters. Compared to TS, VS increases flexibility by allowing different scaling and shifting per class, at the cost of a higher risk of overfitting.
We use implementations from the \texttt{probmetrics} package, inspired by \citet{ranjan2023torchcal}.

\paragraph{Matrix scaling} \citep{guo2017calibration} (MS) further extends vector scaling by applying a full linear transformation to the logits:
\[
\mathrm{MS}(\mathbf{p}) = \mathrm{softmax}(W \mathbf{z} + \mathbf{b}) \; ,
\]
where $W \in \mathbb{R}^{K \times K}$ is a weight matrix and $\mathbf{b} \in \mathbb{R}^K$ is a bias vector.
This formulation can capture interactions between classes but introduces $K(K+1)$ parameters, making it prone to overfitting unless the calibration set is large.
We use implementations from the \texttt{probmetrics} package, inspired by \citet{ranjan2023torchcal}.

\paragraph{Dirichlet calibration} \citep{kull2019beyond} (Dirichlet) is a multiclass calibration method derived from the Dirichlet distribution, which generalizes the Beta calibration method used for binary classification.
Just like MS, it applies multinomial logistic regression directly to the log-transformed class probabilities.
To prevent over-parameterization, Dirichlet calibration employs Off-Diagonal and Intercept Regularization (ODIR) to penalize large off-diagonal weights.
We use the implementation provided in the \texttt{dirichletcal} Python package.

\paragraph{Structured Matrix Scaling} \citep{berta2026structured} (SMS) is another extension of MS that applies the same model but uses a hierarchical regularization structure to penalize intercept, off-diagonal, and diagonal parameters separately. The regularization strength chosen for each parameter group is a function of the number of calibration samples and number of classes.
We use implementations from the \texttt{probmetrics} package.

\paragraph{Structured Vector Scaling} \citep{berta2026structured} (SVS) is a reduction of SMS that uses the same regularization structure but restricted to a diagonal (VS) logistic model.
We use implementations from the \texttt{probmetrics} package.

\paragraph{Kernel-based calibration} with a Dirichlet kernel \citep{popordanoska2022consistent} (Kernel) generalizes the nonparametric Beta kernel approach to $K$-class classification ($K>2$).
To properly account for the geometry of the probability simplex $\Delta_K$ and avoid boundary artifacts, the estimator uses a Dirichlet kernel.
For a new prediction vector $\mathbf{p} \in \Delta_K$, the calibrated probability for class $k$ is obtained through a class-wise Nadaraya-Watson estimator:
\[
\mathrm{Kernel}(\mathbf{p})_k = \frac{\sum_{i=1}^N \mathbb{I}(Y_i = k) K_h(\mathbf{p}, \mathbf{p}_i) }{\sum_{i=1}^N K_h(\mathbf{p}, \mathbf{p}_i)} \; ,
\]
where $\mathbb{I}(\cdot)$ is the indicator function and $K_h(\mathbf{p}, \mathbf{p}_i)$ is the Dirichlet probability density function evaluated at $\mathbf{p}$ with concentration parameters $\alpha = \mathbf{p}_i / h + 1$.
Similar to the binary case, we cap the calibration set at 10,000 samples to ensure scalable, stable pairwise kernel evaluations.
The bandwidth $h$ is set using a simplex-adapted Scott's rule heuristic, $h = N^{-2/(d+4)}$, where the intrinsic dimensionality is $d=K-1$.
The implementation is in the \texttt{probmetrics} package.

\paragraph{Tree-based calibration.} The \textbf{LightGBM} \citep{ke2017lightgbm}, \textbf{XGBoost} \citep{chen2016xgboost} and \textbf{CatBoost} \citep{prokhorenkova2018catboost} calibrators described in our list of binary calibration methods transfer to multiclass calibration straightforwardly. We evaluate them as multiclass calibration methods in our benchmarks, except on the \textbf{ImageNet-multiclass} benchmark, where they are prohibitively slow.

\paragraph{One-versus-rest calibration} (OvR) adapts binary calibration methods to the multiclass setting. For each class $k \in \{ 1, \dots, K \}$, a binary calibration function $g_k : [0,1] \to [0,1]$ is learned to estimate the probability of the event $Y = k$ versus $Y \neq k$, using the original class probability $p_k$ as input. At inference time, each class probability is calibrated independently:
\[
\tilde{p}_k = g_k(p_k) \, .
\]
Since the resulting vector $\tilde{\mathbf{p}}$ does not necessarily sum to one, it is normalized to produce a valid probability distribution:
\[
\mathrm{OvR}(\mathbf{p})_k = \frac{\tilde{p}_k}{\sum_{i=1}^K \tilde{p}_i} \; .
\]
This approach is simple and flexible, allowing the use of any binary calibration method in the multiclass setting. However, it may distort relative class probabilities due to the independent calibration of each component. In our benchmark, we apply the following binary methods in an OvR fashion: histogram binning (uniform and quantile), isotonic regression, centered isotonic regression (CIR), Bayesian binning into quantiles (BBQ), Venn-Abers, spline calibration and Platt scaling on the logits (Platt-logits).
For Platt-logits, each binary calibrator is an affine logistic model on the one-versus-rest log-odds of class $k$,
\[
g_k(p_k) = \sigma\big(\alpha_k \sigma^{-1}(p_k) + \beta_k\big) = \sigma\Big(\alpha_k \log \frac{p_k}{1 - p_k} + \beta_k\Big) \; ,
\]
so that the multiclass model has $2K$ parameters, each pair being fitted independently by minimizing the binary logloss, which makes this method very fast.

While promising attempts have been made to extend nonparametric binary methods to the multiclass setting \citep{berta2024classifier, bao2026brenier}, we cannot include them in our benchmark as-is, because they scale poorly to high-dimensional predictions and fitting is too slow.

\section{Limitations}
\label{app:Limitations}

In this section, we outline some limitations of our current benchmark that could be addressed in a future version.

\paragraph{Investigate the impact of hyperparameters.}
For simplicity, our benchmark evaluates post-hoc calibration methods using fixed hyperparameters.
However, several methods feature tunable parameters that could benefit from cross-validation, such as the regularization strengths in Dirichlet, SVS, and SMS.
Implementing such tuning requires fitting each method multiple times per task, which is currently computationally prohibitive for the slower algorithms within our large-scale benchmark framework.
This highlights the importance of good (fast) implementations for post-hoc calibration methods.

\paragraph{Investigate other forms of multiclass calibration.}
By considering post-hoc improvement in proper scores, we explicitly target the full calibration error after post-hoc calibration, whatever the number of classes considered.
Weaker notions of multiclass calibration exist and are very popular as well, like top-class calibration, which considers only the calibration of the largest probability assigned by the classifier.
Certain calibration methods might excel at these narrower objectives despite performing poorly on full calibration, meaning method rankings could shift under different evaluation metrics.

\paragraph{Larger scale computer vision benchmarks.}
While our tabular benchmarks cover dozens of datasets and models, the diversity and scale of our computer vision evaluations could be further expanded.
Future iterations could incorporate a wider array of modern architectures and diverse image datasets to ensure the robustness of the benchmark's conclusions across vision tasks.

\paragraph{Missing calibration methods.}
Although we sought to include as diverse a set of baseline methods as possible, an exhaustive evaluation of the literature is practically unattainable.
We hope this benchmark encourages authors of current and future post-hoc calibration methods to release open-source, \texttt{scikit-learn}-compatible, and computationally efficient implementations, allowing us to include them in our benchmark and fostering a collective push toward reproducibility and standardized comparisons.

In particular, several recent calibration methods could not be included in this benchmark due to implementation constraints, computational cost, or lack of publicly available code.
This includes, among others, \citep{rahimi2020intra} and \citep{qian2025extending} that are too slow for running on our whole benchmark for now, \citep{wenger2020nonparametric} for which the available implementation is deprecated, \citep{arad2025improving} for which we did not find an implementation online and \citep{lin2023taking}.
While these approaches are promising and representative of recent advances in post-hoc calibration, incorporating them into a unified and efficient evaluation framework remains challenging.

\paragraph{Calibrating on predicted probabilities.}
We fit every calibration method on predicted probabilities, treating the base model as a black box.
When raw logits are available, logit-based methods could benefit from them, as the softmax discards one dimension of the logit space and saturates confident predictions in finite precision.
Our ablation on the computer vision benchmarks (\Cref{app:logits_ablation}) shows that this choice does not penalize TS, VS and MS, but new methods evaluated on CalArena could be more sensitive to it, and logits are not available for our tabular benchmarks.

\paragraph{Exploring other modalities.}
Finally, extending this benchmark to other data modalities represents a critical path forward.
In particular, investigating the calibration of generative models, such as Large Language Models (LLMs) and other modern Natural Language Processing (NLP) systems, remains an open and highly relevant challenge. However, for next-token prediction, the number of classes is so large that only a few methods are applicable.

\section{ImageNet benchmark results}
\label{app:ImageNet_results}

\begin{figure}[htbp]
    \centering
    \includegraphics[width=0.5\linewidth]{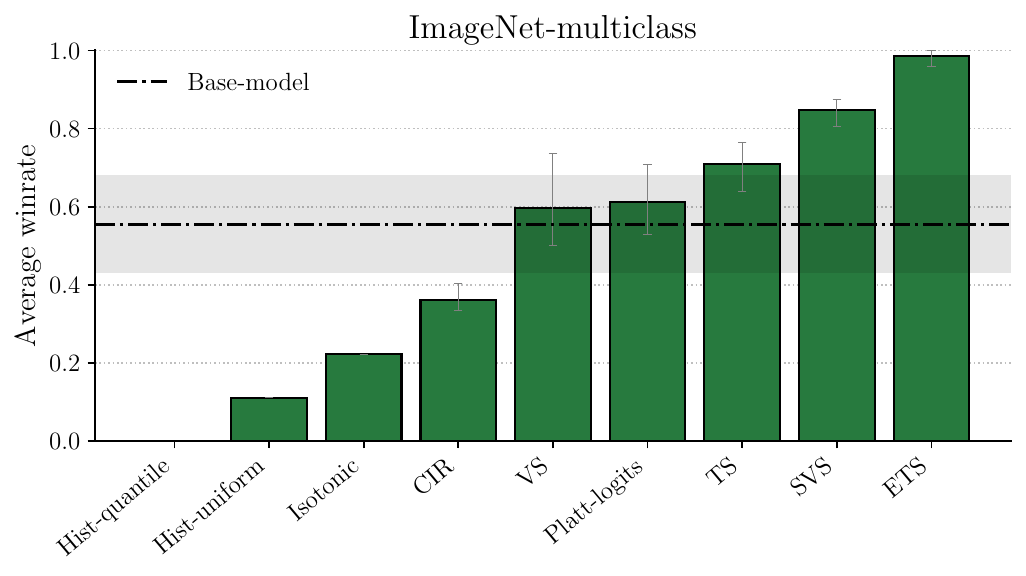}
    \caption{
    Benchmark results for \textbf{ImageNet-multiclass}.
    Each bar represents the winrate of the corresponding method, averaged over all experiments in the benchmark, with 95\% CIs constructed by bootstrapping experiments.
    }
    \label{fig:ImageNetWinratesResults}
\end{figure}

On the \textbf{ImageNet-multiclass} benchmark, containing only ImageNet predictions (1000 classes), several calibration methods cannot be applied. Matrix-scaling type methods (MS, SMS, Dirichlet) would require fitting around a million parameters, which is prohibitively slow.
Binary methods applied OvR need to be fitted 1000 times so only very fast methods can be used; we include Isotonic regression, CIR, the two histogram regressions and Platt-logits.
Every OvR method except Platt-logits degrades the performance of the initial model, and Platt-logits remains behind TS, which has a single parameter, demonstrating the limits of this approach for high-dimensional problems.
ETS ranks first with almost $1.0$ winrate, above SVS, TS, Platt-logits and VS.

\section{Calibrating on probabilities vs.\ logits}
\label{app:logits_ablation}

\paragraph{Design choice.}
In CalArena, every post-hoc calibration method is fitted on the probabilities $\mathbf{p} = f(X)$ predicted by the base model.
Methods that operate on logits, like TS, VS and MS, are applied to the log-probabilities $\mathbf{z} = \log \mathbf{p}$ (see \Cref{app:multiclass_calibration_methods}).
We made this choice because we aim to compare calibration methods that can be applied to any black-box predictor: predicted probabilities are the common interface that every classifier provides, while raw logits are not always accessible.
In particular, \texttt{TabRepo} and \texttt{TabArena} only store predicted probabilities.
Moreover, post-hoc calibration is defined as learning a function $g$ applied to the output $f(X)$ of a frozen model; using internal quantities of $f$ amounts to modifying the model itself rather than correcting it after the fact.

\paragraph{What is lost when calibrating probabilities.}
When raw logits $\mathbf{s} \in \mathbb{R}^K$ are available, with $\mathbf{p} = \mathrm{softmax}(\mathbf{s})$, two effects could in principle make them more informative inputs for logit-based calibration methods:
\begin{enumerate}[topsep=0mm, itemsep=0mm, leftmargin=5mm]
    \item \emph{Dimensional compression.} The softmax is invariant to adding a constant to all logits, so the probabilities lose one dimension of information compared to the raw logits.
    \item \emph{Numerical compression.} In finite precision, very confident predictions saturate: different logits can be mapped to probabilities that are exactly $0$ or $1$ (our benchmarks store probabilities in single precision), which cannot be distinguished anymore after the softmax.
\end{enumerate}
To separate the two effects, we consider \emph{scaled logits} $\mathbf{s} - \log \sum_{j=1}^K \exp(s_j)$, which are mathematically equal to the log-probabilities $\log \mathrm{softmax}(\mathbf{s})$ but computed without going through the softmax.
Scaled logits are thus only affected by dimensional compression, while log-probabilities are affected by both effects.
Note that TS is invariant to a per-sample shift of the logits, $\mathrm{softmax}((\mathbf{s} - c)/T) = \mathrm{softmax}(\mathbf{s}/T)$, so dimensional compression does not affect it, and its raw-logit and scaled-logit versions only differ through numerics.
This is not the case for VS and MS, whose class-specific parameters break this invariance.

\begin{figure}[htbp]
    \centering
    \begin{minipage}{\textwidth}
        \centering
        \includegraphics[width=\linewidth]{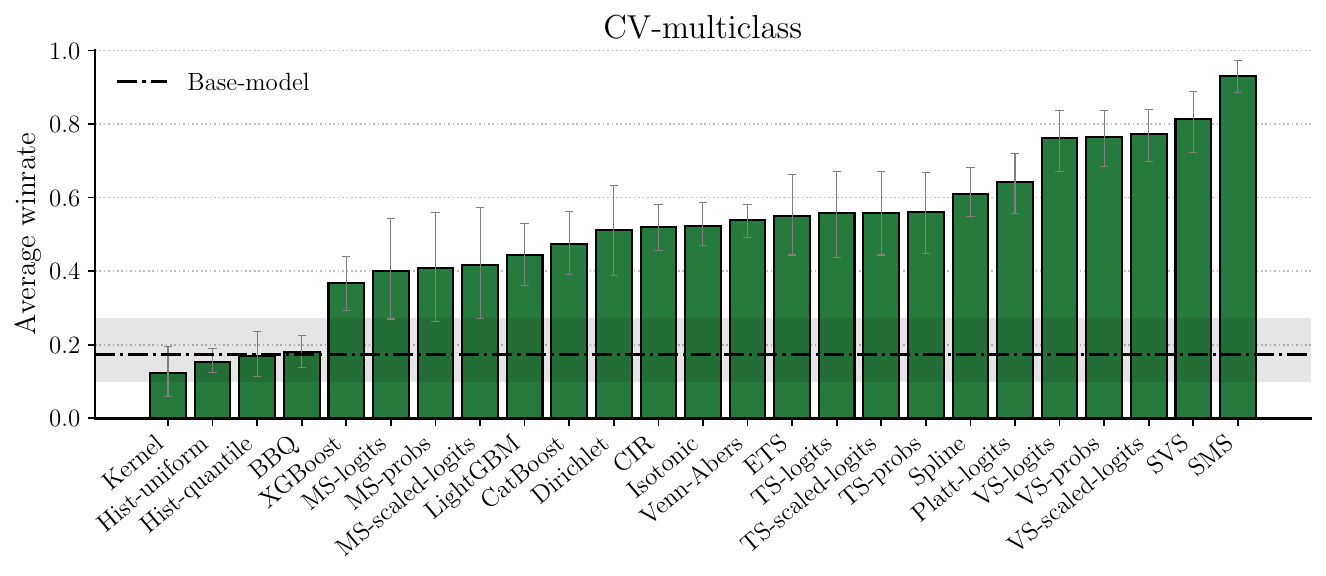}
    \end{minipage}
    \begin{minipage}{\textwidth}
        \centering
        \includegraphics[width=0.75\linewidth]{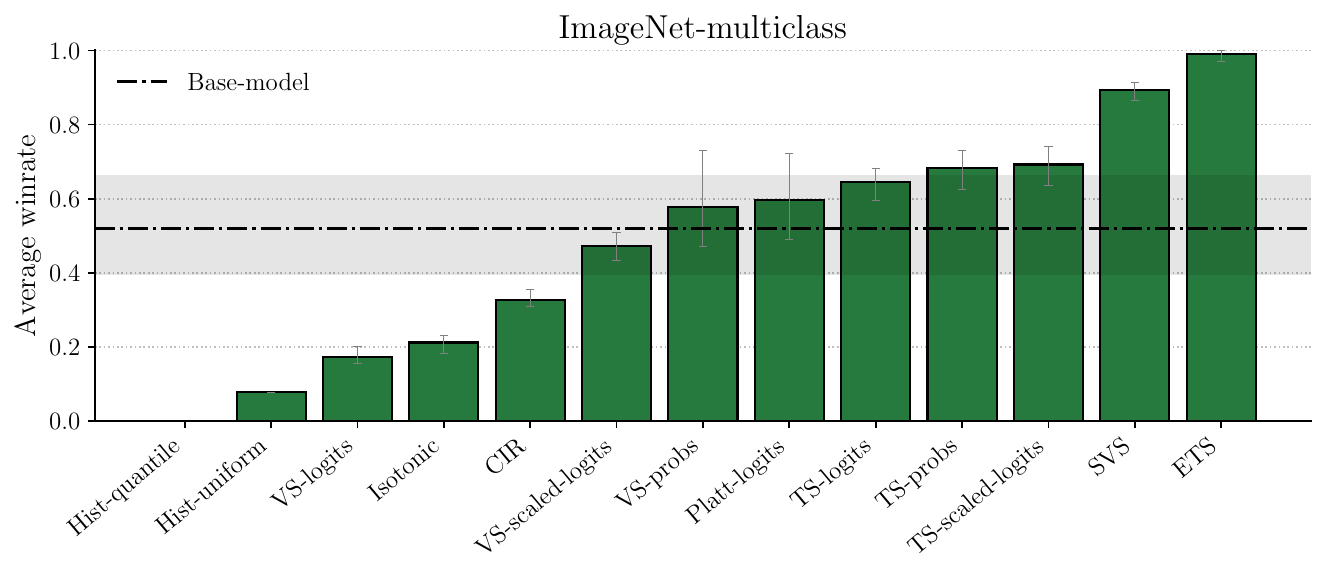}
    \end{minipage}
    \caption{
    Logits vs.\ probabilities ablation on \textbf{CV-multiclass} (top) and \textbf{ImageNet-multiclass} (bottom).
    Logit-based methods are fitted on log-probabilities (\textbf{-probs}, used in the rest of the paper), on logits minus their log-sum-exp (\textbf{-scaled-logits}) or on raw logits (\textbf{-logits}).
    Each bar represents the Brier score winrate of the method, averaged over all experiments in the benchmark, with 95\% CIs constructed by bootstrapping experiments.
    }
    \label{fig:LogitsAblation}
\end{figure}

\paragraph{Experimental setup.}
For the computer vision benchmarks, we have access to the raw logits of the base models.
We compare three versions of the most standard logit-based methods, TS, VS and MS (only TS and VS on \textbf{ImageNet-multiclass}, where MS is too slow):
\begin{itemize}[topsep=0mm, itemsep=0mm, leftmargin=5mm]
    \item \textbf{\{method\}-probs} is fitted on log-probabilities, as in the rest of the paper (dimensional and numerical compression);
    \item \textbf{\{method\}-scaled-logits} is fitted on scaled logits (dimensional compression only);
    \item \textbf{\{method\}-logits} is fitted on the raw logits (no compression).
\end{itemize}
We add these variants to the \textbf{CV-multiclass} and \textbf{ImageNet-multiclass} benchmarks and report the resulting Brier score winrate leaderboards in \Cref{fig:LogitsAblation}.
The code to reproduce this ablation is available in the \texttt{ablation\_logits} folder of our repository.

\paragraph{Results.}
On \textbf{CV-multiclass}, the three variants of each method land within each other's confidence intervals, and their apparent ordering is not consistent across methods: it is not statistically distinguishable from noise.
Crucially, the variants do not change the ranking of TS, VS and MS against the other calibration methods of the benchmark.
For TS, the three variants yield the same Brier scores up to the fifth decimal, as expected from its shift invariance.

On \textbf{ImageNet-multiclass}, TS is again unaffected: the three variants yield identical Brier scores up to the fifth decimal, and their winrate differences only come from the arbitrary resolution of near-ties.
VS fitted on scaled logits performs on par with VS fitted on log-probabilities (the average absolute difference in Brier score is below $10^{-3}$), showing that numerical compression has no measurable effect, even though ImageNet models make very confident predictions.
On the other hand, VS fitted on raw logits performs much worse and never improves over the base model.
On raw logits, the per-class scaling parameters of VS also act on the per-sample offset of the logits, which carries no information on the predicted distribution; with 1000 classes, this extra degree of freedom appears to harm the fit rather than help it.
In this case, the dimensional compression induced by the softmax is beneficial.

To sum up, numerical compression has no measurable effect in our experiments, and dimensional compression is either neutral or beneficial.
Fitting calibration methods on predicted probabilities, as we do throughout the paper, does not penalize logit-based methods on our benchmarks.
Computer vision models make very confident predictions, close to the boundaries of the simplex, so numerical compression should be even less of a concern for our tabular benchmarks, for which logits are not available anyway.
We note, however, that this ablation does not guarantee that the same holds for every new calibration method evaluated with CalArena.

\section{Calibrator runtimes}
\label{app:Runtimes}

In this section we compare calibrator runtimes.
We report the average time elapsed for calibrator fitting and performing predictions on the test set over the TabRepo, TabArena and CV benchmarks.
We normalize runtimes by dividing by the number of calibration samples and multiplying by 1000, to get an average time per 1000 samples.
For multiclass experiments we also divide the runtime by the number of classes to get an average runtime per 1000 samples per class.

We run every experiment on a single Cascade Lake Intel Xeon Gold 5218 CPU with 10 gigabytes of RAM for the binary experiments and 20 gigabytes for the multiclass experiments.
We refer interested readers to our SLURM execution scripts released in the CalArena package for the configuration used to run different methods.

\begin{figure}
    \centering
    \includegraphics[width=\linewidth]{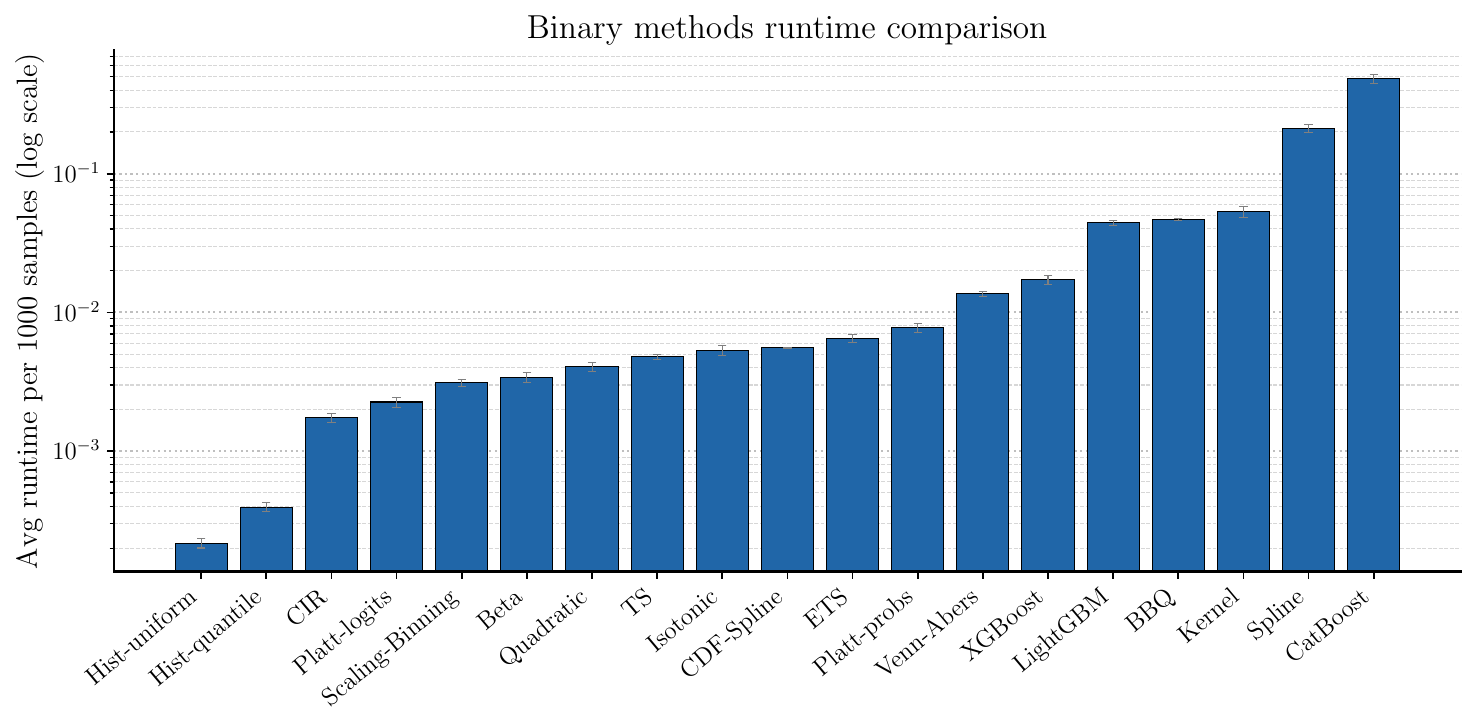}
    \caption{
    Average runtime (seconds, fitting on the calibration set plus predicting on the test set) per 1000 samples in the calibration set for all binary calibrators.
    Averages are taken over all experiments in the TabRepo-binary, TabArena-binary and CV-binary benchmarks.
    }
    \label{fig:binary_runtimes}
\end{figure}

\begin{figure}
    \centering
    \includegraphics[width=\linewidth]{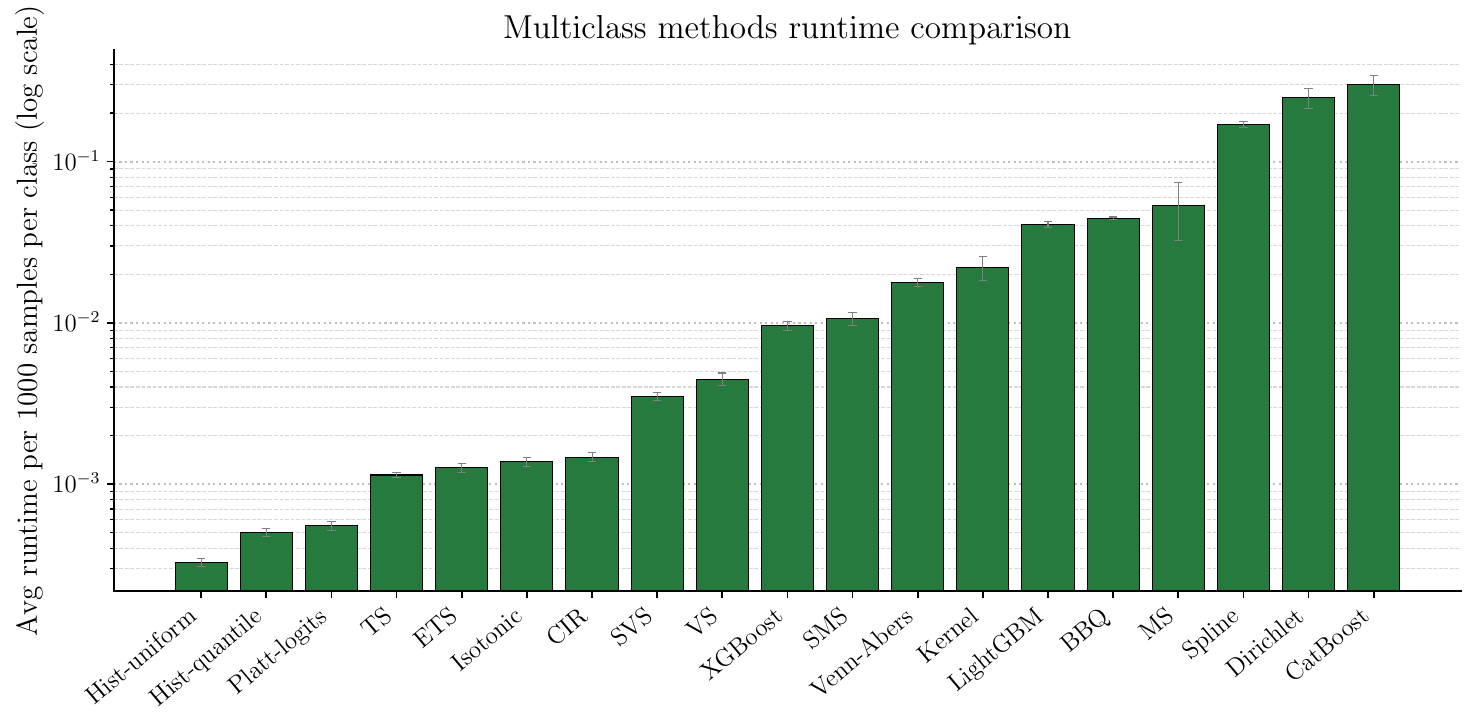}
    \caption{
    Average runtime (seconds, fitting on the calibration set plus predicting on the test set) per 1000 samples in the calibration set per class for all multiclass calibrators.
    Averages are taken over all experiments in the TabRepo-multiclass, TabArena-multiclass and CV-multiclass benchmarks.
    }
    \label{fig:multiclass_runtimes}
\end{figure}

We report the average runtimes for our binary calibrators in \Cref{fig:binary_runtimes} and multiclass calibrators in \Cref{fig:multiclass_runtimes}.

\section{Additional result aggregation methods}
\label{app:aggregation}

Besides the winrates used in the main text, we aggregate benchmark results with the three methods below.

\paragraph{Elo scores in a Bradley-Terry model.}
Following a trend in recent benchmarks \citep{chiang2024chatbot, erickson2025tabarena}, an alternative option is to compute Elo scores for each post-hoc calibration method by treating each experiment as a set of one-vs-one matches where method $i$ beats method $j$ if $s_i > s_j$.
Elo scores for each method are computed using a Bradley-Terry model \citep{bradley1952rank} using the \texttt{arena\_rank} Python package \citep{chiang2024chatbot}.
We compute 95\% CIs on the Elo scores by bootstrapping datasets or experiments directly.
We present leaderboards for our benchmarks in \Cref{app:elo_results}.

\paragraph{Absolute improvements.}
One issue with such rank-based aggregations is that the scale of the improvement is not considered.
If $\Phi_\mathrm{BS}$ for method $i$ is marginally larger than for method $j$, this is still considered a win, and has the same impact on the final ranking as if the improvement is large.
To account for this, we present raw improvement results in \Cref{app:absolute_results}, where we report post-hoc improvements for each method, averaged over all benchmark experiments.
One weakness of this aggregation is that the scale of improvement varies a lot with models and datasets, making classical CIs less informative and giving more influence to experiments for which the initial loss is larger.

\paragraph{Statistical analysis.}
Finally, one can raise the issue of statistical significance.
When comparing several classifiers over multiple experiments (which is what we do by comparing classifiers post-processed with different calibration functions), a standard approach is the evaluation procedure proposed by \citet{demsar2006statistical}.
We present results obtained with this procedure in \Cref{app:statistical_analysis}.
We first reject the null hypothesis that all post-hoc calibration methods are equivalent with a Friedman test, and then perform pairwise comparisons with Nemenyi post-hoc tests \citep{nemenyi1963distribution} to determine which methods are statistically distinguishable.
We use the \texttt{scikit-posthocs} Python package \citep{Terpilowski2019}, and communicate results with Critical Differences (CD) diagrams for each benchmark.

\section{Elo score results}
\label{app:elo_results}

\begin{figure}[htbp]
    \centering
    \begin{minipage}{\textwidth}
        \centering
        \includegraphics[width=\linewidth]{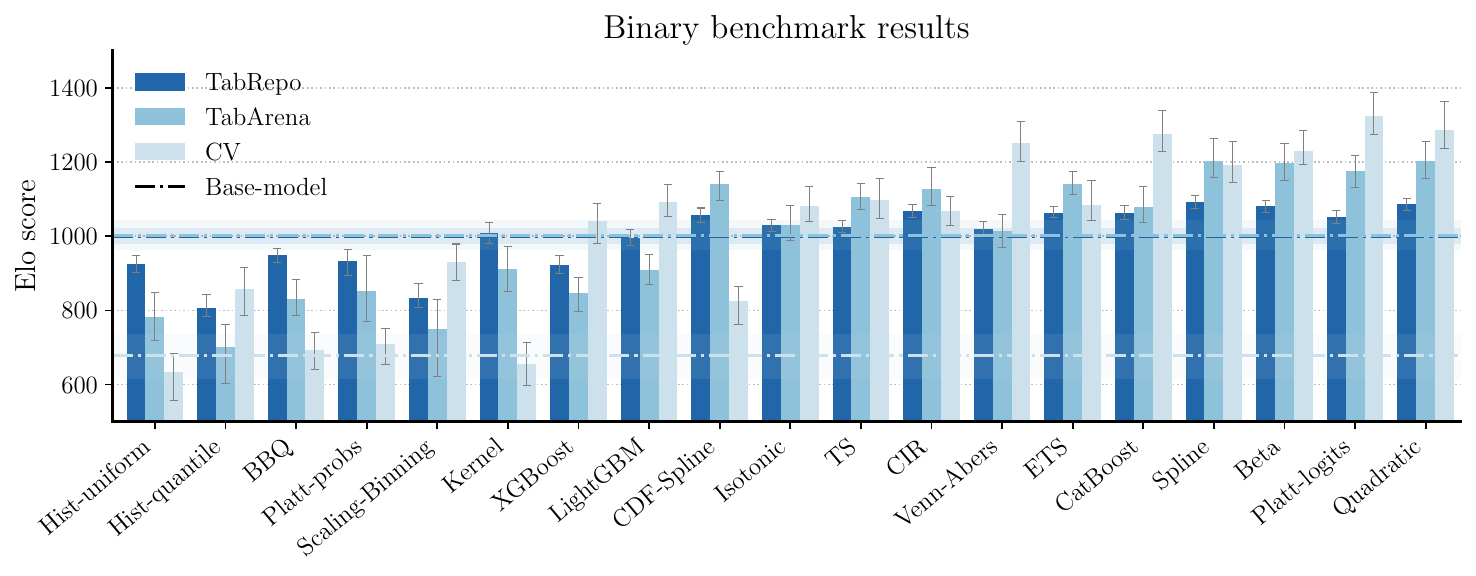}
    \end{minipage}
    \begin{minipage}{\textwidth}
        \centering
        \includegraphics[width=\linewidth]{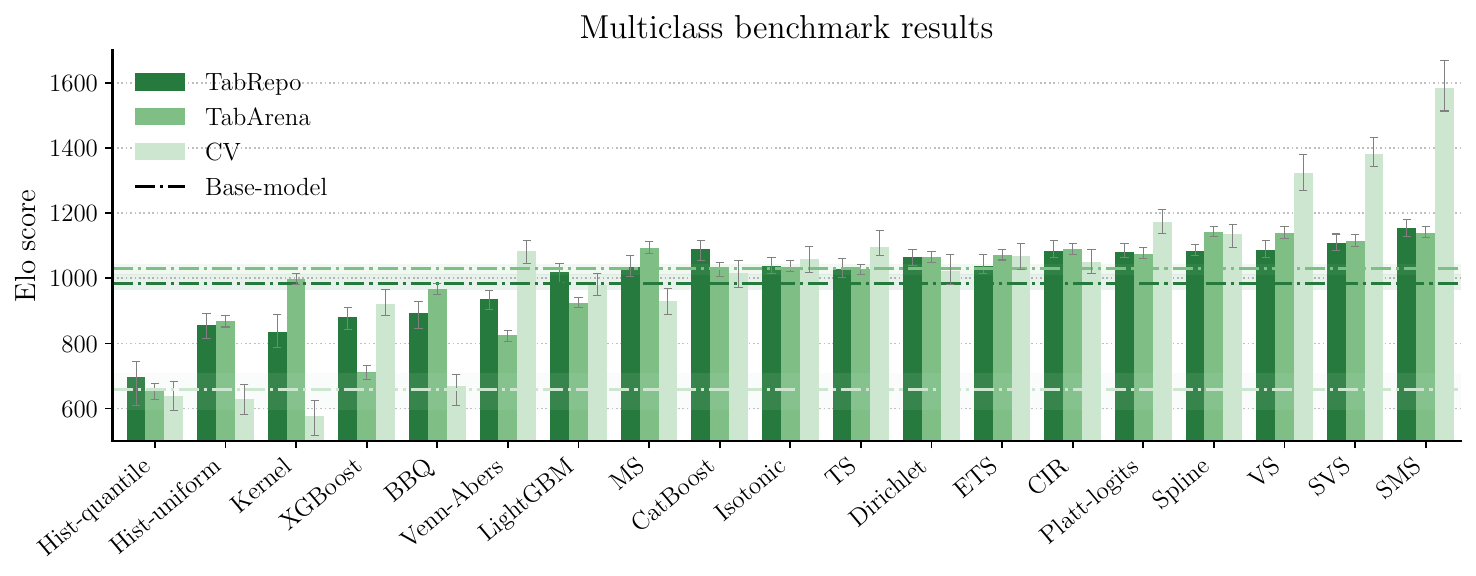}
    \end{minipage}
    \caption{
    Benchmark results for binary post-hoc calibration benchmarks \textbf{TabRepo-binary}, \textbf{TabArena-binary} and \textbf{CV-binary} (top) and multiclass post-hoc calibration benchmarks \textbf{TabRepo-multiclass}, \textbf{TabArena-multiclass} and \textbf{CV-multiclass} (bottom).
    Each bar represents the Elo score of the calibration method, with 95\% CIs constructed by bootstrapping entire datasets (TabRepo and TabArena-binary benchmarks) or experiments directly (TabArena-multiclass and CV benchmarks).
    Methods are ranked based on the average Elo score over the three benchmarks.
    }
    \label{fig:EloResults}
\end{figure}

We provide results using Elo ratings in \Cref{fig:EloResults}.

\FloatBarrier

\section{Absolute results}
\label{app:absolute_results}

We evaluate the absolute post-hoc improvement on different benchmarks in Tables \ref{tab:ImprovementsTabRepoBinary}--\ref{tab:ImprovementsImageNetMulticlass}.

\begin{table}[htbp]
\centering
\caption{
Absolute post-hoc improvement for 5 metrics of interest, averaged over all experiments in the \textbf{TabRepo-binary} post-hoc calibration benchmark.
\textbf{For readability, every value in the table is multiplied by 100.}
The $\pm$ bounds indicate standard 95\% CIs calculated using $n = $ number of unique datasets in the benchmark.
Methods are ranked by PHI in Brier score and we indicate ranks for each metric in parentheses.
}
\label{tab:ImprovementsTabRepoBinary}
\resizebox{\linewidth}{!}{
\begin{tabular}{llllll}
\toprule
Method & $\phantom{-} 100 \times \Phi_\mathrm{Brier}$ & $\phantom{-} 100 \times \Phi_\mathrm{Logloss}$ & $\phantom{-} 100 \times \Phi_\mathrm{Kuiper}$ & $\phantom{-} 100 \times \Phi_\mathrm{ECE-15}$ & $\phantom{-} 100 \times \Phi_\mathrm{Accuracy}$ \\
\midrule
Quadratic & $\phantom{-}0.36$ {\tiny $\pm 0.39$} (\#1) & $\phantom{-}1.27$ {\tiny $\pm 2.44$} (\#6) & $\phantom{-}0.43$ {\tiny $\pm 0.31$} (\#5) & $\phantom{-}0.92$ {\tiny $\pm 0.62$} (\#13) & $\phantom{-}0.03$ {\tiny $\pm 0.26$} (\#4) \\
Spline & $\phantom{-}0.35$ {\tiny $\pm 0.39$} (\#2) & $\phantom{-}0.86$ {\tiny $\pm 2.92$} (\#13) & $\phantom{-}0.44$ {\tiny $\pm 0.31$} (\#3) & $\phantom{-}1.03$ {\tiny $\pm 0.64$} (\#11) & $-0.01$ {\tiny $\pm 0.32$} (\#9) \\
ETS & $\phantom{-}0.35$ {\tiny $\pm 0.38$} (\#3) & $\phantom{-}1.63$ {\tiny $\pm 2.33$} (\#3) & $\phantom{-}0.14$ {\tiny $\pm 0.47$} (\#12) & $\phantom{-}0.78$ {\tiny $\pm 0.62$} (\#16) & $\phantom{-}0.01$ {\tiny $\pm 0.07$} (\#6) \\
Beta & $\phantom{-}0.34$ {\tiny $\pm 0.39$} (\#4) & $\phantom{-}1.22$ {\tiny $\pm 2.46$} (\#8) & $\phantom{-}0.43$ {\tiny $\pm 0.31$} (\#4) & $\phantom{-}0.97$ {\tiny $\pm 0.62$} (\#12) & $\phantom{-}0.02$ {\tiny $\pm 0.27$} (\#5) \\
CIR & $\phantom{-}0.30$ {\tiny $\pm 0.40$} (\#5) & $-16.19$ {\tiny $\pm 16.62$} (\#19) & $\phantom{-}0.45$ {\tiny $\pm 0.31$} (\#2) & $\phantom{-}1.10$ {\tiny $\pm 0.63$} (\#10) & $-0.02$ {\tiny $\pm 0.31$} (\#11) \\
Platt-logits & $\phantom{-}0.30$ {\tiny $\pm 0.39$} (\#6) & $\phantom{-}1.11$ {\tiny $\pm 2.37$} (\#9) & $\phantom{-}0.35$ {\tiny $\pm 0.31$} (\#8) & $\phantom{-}0.75$ {\tiny $\pm 0.64$} (\#17) & $\phantom{-}0.04$ {\tiny $\pm 0.25$} (\#3) \\
CDF-Spline & $\phantom{-}0.30$ {\tiny $\pm 0.39$} (\#7) & $\phantom{-}1.78$ {\tiny $\pm 3.39$} (\#1) & $\phantom{-}0.41$ {\tiny $\pm 0.31$} (\#6) & $\phantom{-}0.85$ {\tiny $\pm 0.59$} (\#14) & $-0.02$ {\tiny $\pm 0.30$} (\#10) \\
CatBoost & $\phantom{-}0.30$ {\tiny $\pm 0.42$} (\#8) & $\phantom{-}1.66$ {\tiny $\pm 2.36$} (\#2) & $\phantom{-}0.45$ {\tiny $\pm 0.33$} (\#1) & $\phantom{-}1.29$ {\tiny $\pm 0.66$} (\#8) & $-0.11$ {\tiny $\pm 0.41$} (\#14) \\
Venn-Abers & $\phantom{-}0.28$ {\tiny $\pm 0.40$} (\#9) & $\phantom{-}1.62$ {\tiny $\pm 2.35$} (\#4) & $-0.11$ {\tiny $\pm 0.47$} (\#17) & $\phantom{-}1.41$ {\tiny $\pm 0.68$} (\#7) & $-0.06$ {\tiny $\pm 0.32$} (\#13) \\
Kernel & $\phantom{-}0.26$ {\tiny $\pm 0.36$} (\#10) & $\phantom{-}0.94$ {\tiny $\pm 2.44$} (\#12) & $\phantom{-}0.26$ {\tiny $\pm 0.30$} (\#9) & $\phantom{-}1.21$ {\tiny $\pm 0.60$} (\#9) & $\phantom{-}0.07$ {\tiny $\pm 0.28$} (\#2) \\
TS & $\phantom{-}0.24$ {\tiny $\pm 0.39$} (\#11) & $\phantom{-}1.26$ {\tiny $\pm 2.24$} (\#7) & $\phantom{-}0.01$ {\tiny $\pm 0.46$} (\#13) & $\phantom{-}0.67$ {\tiny $\pm 0.63$} (\#18) & $-0.01$ {\tiny $\pm 0.08$} (\#8) \\
Isotonic & $\phantom{-}0.23$ {\tiny $\pm 0.43$} (\#12) & $-16.22$ {\tiny $\pm 16.66$} (\#20) & $-0.07$ {\tiny $\pm 0.49$} (\#16) & $\phantom{-}1.81$ {\tiny $\pm 0.68$} (\#5) & $-0.06$ {\tiny $\pm 0.33$} (\#12) \\
Platt-probs & $\phantom{-}0.16$ {\tiny $\pm 0.39$} (\#13) & $\phantom{-}1.02$ {\tiny $\pm 2.38$} (\#10) & $\phantom{-}0.18$ {\tiny $\pm 0.33$} (\#10) & $\phantom{-}0.60$ {\tiny $\pm 0.65$} (\#19) & $\phantom{-}0.07$ {\tiny $\pm 0.23$} (\#1) \\
LightGBM & $\phantom{-}0.16$ {\tiny $\pm 0.43$} (\#14) & $\phantom{-}1.43$ {\tiny $\pm 2.37$} (\#5) & $\phantom{-}0.38$ {\tiny $\pm 0.34$} (\#7) & $\phantom{-}1.43$ {\tiny $\pm 0.70$} (\#6) & $-0.16$ {\tiny $\pm 0.39$} (\#16) \\
BBQ & $\phantom{-}0.10$ {\tiny $\pm 0.42$} (\#15) & $\phantom{-}0.37$ {\tiny $\pm 2.61$} (\#15) & $-0.03$ {\tiny $\pm 0.43$} (\#15) & $\phantom{-}2.42$ {\tiny $\pm 0.76$} (\#2) & $-0.18$ {\tiny $\pm 0.36$} (\#17) \\
Hist-uniform & $\phantom{-}0.08$ {\tiny $\pm 0.43$} (\#16) & $-10.41$ {\tiny $\pm 17.05$} (\#17) & $-0.28$ {\tiny $\pm 0.46$} (\#20) & $\phantom{-}2.13$ {\tiny $\pm 0.68$} (\#4) & $-0.13$ {\tiny $\pm 0.36$} (\#15) \\
XGBoost & $\phantom{-}0.06$ {\tiny $\pm 0.44$} (\#17) & $\phantom{-}1.02$ {\tiny $\pm 2.38$} (\#11) & $\phantom{-}0.18$ {\tiny $\pm 0.34$} (\#11) & $\phantom{-}0.79$ {\tiny $\pm 0.68$} (\#15) & $-0.27$ {\tiny $\pm 0.42$} (\#18) \\
Base-model & $\phantom{-}0.00$ {\tiny $\pm 0.00$} (\#18) & $\phantom{-}0.00$ {\tiny $\pm 0.00$} (\#16) & $\phantom{-}0.00$ {\tiny $\pm 0.00$} (\#14) & $\phantom{-}0.00$ {\tiny $\pm 0.00$} (\#20) & $\phantom{-}0.00$ {\tiny $\pm 0.00$} (\#7) \\
Scaling-Binning & $-0.32$ {\tiny $\pm 0.45$} (\#19) & $\phantom{-}0.65$ {\tiny $\pm 2.36$} (\#14) & $-0.23$ {\tiny $\pm 0.36$} (\#18) & $\phantom{-}2.26$ {\tiny $\pm 0.69$} (\#3) & $-0.36$ {\tiny $\pm 0.37$} (\#19) \\
Hist-quantile & $-0.44$ {\tiny $\pm 0.49$} (\#20) & $-12.52$ {\tiny $\pm 16.01$} (\#18) & $-0.26$ {\tiny $\pm 0.37$} (\#19) & $\phantom{-}2.44$ {\tiny $\pm 0.72$} (\#1) & $-0.57$ {\tiny $\pm 0.48$} (\#20) \\
\bottomrule
\end{tabular}

}
\end{table}

\begin{table}[htbp]
\centering
\caption{
Absolute post-hoc improvement for 5 metrics of interest, averaged over all experiments in the \textbf{TabArena-binary} post-hoc calibration benchmark.
\textbf{For readability, every value in the table is multiplied by 100.}
The $\pm$ bounds indicate standard 95\% CIs calculated using $n = $ number of unique datasets in the benchmark.
Methods are ranked by PHI in Brier score and we indicate ranks for each metric in parentheses.
}
\label{tab:ImprovementsTabArenaBinary}
\resizebox{\linewidth}{!}{
\begin{tabular}{llllll}
\toprule
Method & $\phantom{-} 100 \times \Phi_\mathrm{Brier}$ & $\phantom{-} 100 \times \Phi_\mathrm{Logloss}$ & $\phantom{-} 100 \times \Phi_\mathrm{Kuiper}$ & $\phantom{-} 100 \times \Phi_\mathrm{ECE-15}$ & $\phantom{-} 100 \times \Phi_\mathrm{Accuracy}$ \\
\midrule
Quadratic & $\phantom{-}0.25$ {\tiny $\pm 0.47$} (\#1) & $\phantom{-}1.43$ {\tiny $\pm 2.46$} (\#1) & $\phantom{-}0.83$ {\tiny $\pm 0.77$} (\#1) & $\phantom{-}0.89$ {\tiny $\pm 0.81$} (\#7) & $\phantom{-}0.11$ {\tiny $\pm 0.33$} (\#3) \\
Platt-logits & $\phantom{-}0.25$ {\tiny $\pm 0.48$} (\#2) & $\phantom{-}1.39$ {\tiny $\pm 2.36$} (\#3) & $\phantom{-}0.82$ {\tiny $\pm 0.78$} (\#3) & $\phantom{-}0.85$ {\tiny $\pm 0.80$} (\#11) & $\phantom{-}0.09$ {\tiny $\pm 0.31$} (\#5) \\
Beta & $\phantom{-}0.25$ {\tiny $\pm 0.48$} (\#3) & $\phantom{-}1.43$ {\tiny $\pm 2.48$} (\#2) & $\phantom{-}0.82$ {\tiny $\pm 0.77$} (\#2) & $\phantom{-}0.88$ {\tiny $\pm 0.81$} (\#8) & $\phantom{-}0.11$ {\tiny $\pm 0.33$} (\#1) \\
CDF-Spline & $\phantom{-}0.23$ {\tiny $\pm 0.41$} (\#4) & $\phantom{-}1.18$ {\tiny $\pm 2.22$} (\#9) & $\phantom{-}0.78$ {\tiny $\pm 0.74$} (\#6) & $\phantom{-}0.82$ {\tiny $\pm 0.75$} (\#14) & $\phantom{-}0.08$ {\tiny $\pm 0.26$} (\#7) \\
Spline & $\phantom{-}0.22$ {\tiny $\pm 0.49$} (\#5) & $\phantom{-}1.35$ {\tiny $\pm 2.50$} (\#4) & $\phantom{-}0.80$ {\tiny $\pm 0.79$} (\#5) & $\phantom{-}0.83$ {\tiny $\pm 0.82$} (\#13) & $\phantom{-}0.11$ {\tiny $\pm 0.36$} (\#4) \\
CIR & $\phantom{-}0.21$ {\tiny $\pm 0.48$} (\#6) & $-3.02$ {\tiny $\pm 4.17$} (\#19) & $\phantom{-}0.81$ {\tiny $\pm 0.78$} (\#4) & $\phantom{-}0.85$ {\tiny $\pm 0.79$} (\#10) & $\phantom{-}0.08$ {\tiny $\pm 0.36$} (\#6) \\
CatBoost & $\phantom{-}0.18$ {\tiny $\pm 0.48$} (\#7) & $\phantom{-}1.34$ {\tiny $\pm 2.50$} (\#5) & $\phantom{-}0.77$ {\tiny $\pm 0.78$} (\#7) & $\phantom{-}0.88$ {\tiny $\pm 0.82$} (\#9) & $\phantom{-}0.11$ {\tiny $\pm 0.36$} (\#2) \\
ETS & $\phantom{-}0.17$ {\tiny $\pm 0.24$} (\#8) & $\phantom{-}1.29$ {\tiny $\pm 2.41$} (\#6) & $\phantom{-}0.63$ {\tiny $\pm 0.55$} (\#9) & $\phantom{-}0.65$ {\tiny $\pm 0.61$} (\#15) & $\phantom{-}0.00$ {\tiny $\pm 0.00$} (\#14) \\
Venn-Abers & $\phantom{-}0.16$ {\tiny $\pm 0.48$} (\#9) & $\phantom{-}1.28$ {\tiny $\pm 2.50$} (\#7) & $\phantom{-}0.44$ {\tiny $\pm 0.81$} (\#12) & $\phantom{-}0.89$ {\tiny $\pm 0.83$} (\#6) & $\phantom{-}0.06$ {\tiny $\pm 0.38$} (\#8) \\
Isotonic & $\phantom{-}0.15$ {\tiny $\pm 0.49$} (\#10) & $-2.91$ {\tiny $\pm 4.30$} (\#18) & $\phantom{-}0.41$ {\tiny $\pm 0.82$} (\#13) & $\phantom{-}1.00$ {\tiny $\pm 0.82$} (\#5) & $\phantom{-}0.05$ {\tiny $\pm 0.38$} (\#9) \\
TS & $\phantom{-}0.14$ {\tiny $\pm 0.24$} (\#11) & $\phantom{-}1.23$ {\tiny $\pm 2.29$} (\#8) & $\phantom{-}0.58$ {\tiny $\pm 0.55$} (\#10) & $\phantom{-}0.63$ {\tiny $\pm 0.60$} (\#16) & $\phantom{-}0.00$ {\tiny $\pm 0.00$} (\#14) \\
Kernel & $\phantom{-}0.11$ {\tiny $\pm 0.46$} (\#12) & $\phantom{-}1.06$ {\tiny $\pm 2.40$} (\#11) & $\phantom{-}0.13$ {\tiny $\pm 0.75$} (\#15) & $\phantom{-}0.49$ {\tiny $\pm 0.81$} (\#18) & $\phantom{-}0.04$ {\tiny $\pm 0.36$} (\#11) \\
LightGBM & $\phantom{-}0.05$ {\tiny $\pm 0.50$} (\#13) & $\phantom{-}1.10$ {\tiny $\pm 2.52$} (\#10) & $\phantom{-}0.75$ {\tiny $\pm 0.79$} (\#8) & $\phantom{-}0.84$ {\tiny $\pm 0.88$} (\#12) & $-0.01$ {\tiny $\pm 0.38$} (\#17) \\
Platt-probs & $\phantom{-}0.05$ {\tiny $\pm 0.46$} (\#14) & $\phantom{-}0.74$ {\tiny $\pm 2.43$} (\#13) & $\phantom{-}0.24$ {\tiny $\pm 0.77$} (\#14) & $\phantom{-}0.14$ {\tiny $\pm 0.88$} (\#19) & $\phantom{-}0.04$ {\tiny $\pm 0.34$} (\#10) \\
BBQ & $\phantom{-}0.02$ {\tiny $\pm 0.47$} (\#15) & $\phantom{-}0.71$ {\tiny $\pm 2.37$} (\#14) & $\phantom{-}0.11$ {\tiny $\pm 0.76$} (\#16) & $\phantom{-}1.29$ {\tiny $\pm 0.89$} (\#2) & $\phantom{-}0.00$ {\tiny $\pm 0.35$} (\#13) \\
XGBoost & $\phantom{-}0.01$ {\tiny $\pm 0.51$} (\#16) & $\phantom{-}0.89$ {\tiny $\pm 2.52$} (\#12) & $\phantom{-}0.57$ {\tiny $\pm 0.80$} (\#11) & $\phantom{-}0.58$ {\tiny $\pm 0.88$} (\#17) & $-0.02$ {\tiny $\pm 0.37$} (\#18) \\
Base-model & $\phantom{-}0.00$ {\tiny $\pm 0.00$} (\#17) & $\phantom{-}0.00$ {\tiny $\pm 0.00$} (\#17) & $\phantom{-}0.00$ {\tiny $\pm 0.00$} (\#17) & $\phantom{-}0.00$ {\tiny $\pm 0.00$} (\#20) & $\phantom{-}0.00$ {\tiny $\pm 0.00$} (\#14) \\
Hist-uniform & $-0.02$ {\tiny $\pm 0.47$} (\#18) & $\phantom{-}0.31$ {\tiny $\pm 2.43$} (\#16) & $-0.89$ {\tiny $\pm 1.00$} (\#20) & $\phantom{-}1.22$ {\tiny $\pm 0.86$} (\#4) & $\phantom{-}0.02$ {\tiny $\pm 0.36$} (\#12) \\
Scaling-Binning & $-0.25$ {\tiny $\pm 0.49$} (\#19) & $\phantom{-}0.70$ {\tiny $\pm 2.50$} (\#15) & $-0.26$ {\tiny $\pm 0.81$} (\#18) & $\phantom{-}1.31$ {\tiny $\pm 0.83$} (\#1) & $-0.26$ {\tiny $\pm 0.34$} (\#20) \\
Hist-quantile & $-0.30$ {\tiny $\pm 0.50$} (\#20) & $-3.04$ {\tiny $\pm 6.21$} (\#20) & $-0.28$ {\tiny $\pm 0.82$} (\#19) & $\phantom{-}1.27$ {\tiny $\pm 0.86$} (\#3) & $-0.23$ {\tiny $\pm 0.38$} (\#19) \\
\bottomrule
\end{tabular}

}
\end{table}

\begin{table}[htbp]
\centering
\caption{
Absolute post-hoc improvement for 5 metrics of interest, averaged over all experiments in the \textbf{CV-binary} post-hoc calibration benchmark.
\textbf{For readability, every value in the table is multiplied by 100.}
The $\pm$ bounds indicate standard 95\% CIs calculated using $n = $ number of experiments in the benchmark.
Methods are ranked by PHI in Brier score and we indicate ranks for each metric in parentheses.
}
\label{tab:ImprovementsCVBinary}
\resizebox{\linewidth}{!}{
\begin{tabular}{llllll}
\toprule
Method & $\phantom{-} 100 \times \Phi_\mathrm{Brier}$ & $\phantom{-} 100 \times \Phi_\mathrm{Logloss}$ & $\phantom{-} 100 \times \Phi_\mathrm{Kuiper}$ & $\phantom{-} 100 \times \Phi_\mathrm{ECE-15}$ & $\phantom{-} 100 \times \Phi_\mathrm{Accuracy}$ \\
\midrule
Venn-Abers & $\phantom{-}3.48$ {\tiny $\pm 1.65$} (\#1) & $\phantom{-}115.15$ {\tiny $\pm 61.72$} (\#1) & $\phantom{-}2.34$ {\tiny $\pm 1.60$} (\#4) & $\phantom{-}3.02$ {\tiny $\pm 2.25$} (\#9) & $\phantom{-}0.13$ {\tiny $\pm 0.88$} (\#8) \\
Platt-logits & $\phantom{-}3.39$ {\tiny $\pm 1.53$} (\#2) & $\phantom{-}102.37$ {\tiny $\pm 56.64$} (\#10) & $\phantom{-}2.40$ {\tiny $\pm 1.42$} (\#2) & $\phantom{-}2.77$ {\tiny $\pm 1.82$} (\#13) & $\phantom{-}0.30$ {\tiny $\pm 0.38$} (\#4) \\
CatBoost & $\phantom{-}3.28$ {\tiny $\pm 1.53$} (\#3) & $\phantom{-}114.09$ {\tiny $\pm 61.34$} (\#3) & $\phantom{-}2.34$ {\tiny $\pm 1.35$} (\#5) & $\phantom{-}3.37$ {\tiny $\pm 2.37$} (\#5) & $-0.18$ {\tiny $\pm 1.05$} (\#19) \\
Quadratic & $\phantom{-}3.21$ {\tiny $\pm 1.46$} (\#4) & $\phantom{-}97.11$ {\tiny $\pm 59.64$} (\#12) & $\phantom{-}2.32$ {\tiny $\pm 1.44$} (\#7) & $\phantom{-}2.85$ {\tiny $\pm 1.83$} (\#12) & $\phantom{-}0.03$ {\tiny $\pm 0.55$} (\#12) \\
XGBoost & $\phantom{-}3.18$ {\tiny $\pm 1.50$} (\#5) & $\phantom{-}114.26$ {\tiny $\pm 62.04$} (\#2) & $\phantom{-}2.19$ {\tiny $\pm 1.62$} (\#8) & $\phantom{-}3.05$ {\tiny $\pm 2.70$} (\#8) & $\phantom{-}0.08$ {\tiny $\pm 0.89$} (\#10) \\
Beta & $\phantom{-}3.17$ {\tiny $\pm 1.50$} (\#6) & $\phantom{-}111.55$ {\tiny $\pm 61.33$} (\#6) & $\phantom{-}2.33$ {\tiny $\pm 1.28$} (\#6) & $\phantom{-}3.25$ {\tiny $\pm 2.19$} (\#6) & $\phantom{-}0.55$ {\tiny $\pm 0.49$} (\#2) \\
LightGBM & $\phantom{-}3.06$ {\tiny $\pm 1.42$} (\#7) & $\phantom{-}113.75$ {\tiny $\pm 61.28$} (\#4) & $\phantom{-}2.48$ {\tiny $\pm 1.47$} (\#1) & $\phantom{-}3.63$ {\tiny $\pm 2.43$} (\#2) & $-0.15$ {\tiny $\pm 0.83$} (\#18) \\
Isotonic & $\phantom{-}3.02$ {\tiny $\pm 1.37$} (\#8) & $-23.39$ {\tiny $\pm 89.15$} (\#18) & $\phantom{-}2.13$ {\tiny $\pm 1.23$} (\#9) & $\phantom{-}2.87$ {\tiny $\pm 1.93$} (\#11) & $\phantom{-}0.14$ {\tiny $\pm 0.82$} (\#7) \\
Spline & $\phantom{-}2.85$ {\tiny $\pm 1.33$} (\#9) & $\phantom{-}74.12$ {\tiny $\pm 69.38$} (\#15) & $\phantom{-}2.01$ {\tiny $\pm 1.37$} (\#11) & $\phantom{-}2.74$ {\tiny $\pm 1.91$} (\#14) & $\phantom{-}0.01$ {\tiny $\pm 0.83$} (\#14) \\
Scaling-Binning & $\phantom{-}2.84$ {\tiny $\pm 1.62$} (\#10) & $\phantom{-}112.37$ {\tiny $\pm 61.51$} (\#5) & $\phantom{-}1.91$ {\tiny $\pm 1.67$} (\#13) & $\phantom{-}3.14$ {\tiny $\pm 2.55$} (\#7) & $\phantom{-}0.41$ {\tiny $\pm 2.22$} (\#3) \\
ETS & $\phantom{-}2.82$ {\tiny $\pm 1.40$} (\#11) & $\phantom{-}107.32$ {\tiny $\pm 59.53$} (\#8) & $\phantom{-}1.81$ {\tiny $\pm 1.47$} (\#15) & $\phantom{-}2.33$ {\tiny $\pm 1.82$} (\#16) & $\phantom{-}0.00$ {\tiny $\pm 0.00$} (\#15) \\
TS & $\phantom{-}2.81$ {\tiny $\pm 1.45$} (\#12) & $\phantom{-}102.49$ {\tiny $\pm 56.46$} (\#9) & $\phantom{-}1.72$ {\tiny $\pm 1.41$} (\#16) & $\phantom{-}2.24$ {\tiny $\pm 1.84$} (\#17) & $\phantom{-}0.00$ {\tiny $\pm 0.00$} (\#15) \\
CIR & $\phantom{-}2.60$ {\tiny $\pm 1.19$} (\#13) & $-37.54$ {\tiny $\pm 93.22$} (\#20) & $\phantom{-}1.67$ {\tiny $\pm 0.83$} (\#17) & $\phantom{-}2.16$ {\tiny $\pm 1.23$} (\#18) & $\phantom{-}0.20$ {\tiny $\pm 0.56$} (\#5) \\
Hist-quantile & $\phantom{-}2.60$ {\tiny $\pm 1.42$} (\#14) & $-31.43$ {\tiny $\pm 103.86$} (\#19) & $\phantom{-}1.85$ {\tiny $\pm 1.32$} (\#14) & $\phantom{-}2.44$ {\tiny $\pm 1.81$} (\#15) & $\phantom{-}0.69$ {\tiny $\pm 1.65$} (\#1) \\
CDF-Spline & $\phantom{-}1.73$ {\tiny $\pm 1.02$} (\#15) & $\phantom{-}84.17$ {\tiny $\pm nan$} (\#14) & $\phantom{-}1.99$ {\tiny $\pm 1.38$} (\#12) & $\phantom{-}2.99$ {\tiny $\pm 2.25$} (\#10) & $\phantom{-}0.14$ {\tiny $\pm 0.20$} (\#6) \\
Platt-probs & $\phantom{-}1.08$ {\tiny $\pm 1.15$} (\#16) & $\phantom{-}109.52$ {\tiny $\pm 59.91$} (\#7) & $\phantom{-}1.58$ {\tiny $\pm 1.17$} (\#18) & $\phantom{-}3.59$ {\tiny $\pm 2.71$} (\#3) & $\phantom{-}0.09$ {\tiny $\pm 0.27$} (\#9) \\
BBQ & $\phantom{-}0.69$ {\tiny $\pm 1.04$} (\#17) & $\phantom{-}96.03$ {\tiny $\pm 67.53$} (\#13) & $\phantom{-}2.37$ {\tiny $\pm 1.53$} (\#3) & $\phantom{-}3.95$ {\tiny $\pm 2.91$} (\#1) & $\phantom{-}0.02$ {\tiny $\pm 0.18$} (\#13) \\
Kernel & $\phantom{-}0.60$ {\tiny $\pm 1.08$} (\#18) & $\phantom{-}100.59$ {\tiny $\pm 62.00$} (\#11) & $\phantom{-}1.06$ {\tiny $\pm 0.97$} (\#19) & $\phantom{-}2.11$ {\tiny $\pm 1.92$} (\#19) & $\phantom{-}0.06$ {\tiny $\pm 0.28$} (\#11) \\
Hist-uniform & $\phantom{-}0.36$ {\tiny $\pm 1.23$} (\#19) & $\phantom{-}28.53$ {\tiny $\pm 56.97$} (\#16) & $\phantom{-}2.02$ {\tiny $\pm 1.46$} (\#10) & $\phantom{-}3.42$ {\tiny $\pm 2.80$} (\#4) & $-0.20$ {\tiny $\pm 0.33$} (\#20) \\
Base-model & $\phantom{-}0.00$ {\tiny $\pm 0.00$} (\#20) & $\phantom{-}0.00$ {\tiny $\pm 0.00$} (\#17) & $\phantom{-}0.00$ {\tiny $\pm 0.00$} (\#20) & $\phantom{-}0.00$ {\tiny $\pm 0.00$} (\#20) & $\phantom{-}0.00$ {\tiny $\pm 0.00$} (\#15) \\
\bottomrule
\end{tabular}

}
\end{table}

\begin{table}[htbp]
\centering
\caption{
Absolute post-hoc improvement for 4 metrics of interest, averaged over all experiments in the \textbf{TabRepo-multiclass} post-hoc calibration benchmark.
\textbf{For readability, every value in the table is multiplied by 100.}
The $\pm$ bounds indicate standard 95\% CIs calculated using $n = $ number of unique datasets in the benchmark.
Methods are ranked by PHI in Brier score and we indicate ranks for each metric in parentheses.
}
\label{tab:ImprovementsTabRepoMulticlass}
\resizebox{\linewidth}{!}{
\begin{tabular}{lllll}
\toprule
Method & $\phantom{-} 100 \times \Phi_\mathrm{Brier}$ & $\phantom{-} 100 \times \Phi_\mathrm{Logloss}$ & $\phantom{-} 100 \times \Phi_\mathrm{ECE-15}$ & $\phantom{-} 100 \times \Phi_\mathrm{Accuracy}$ \\
\midrule
CatBoost & $\phantom{-}0.65$ {\tiny $\pm 0.64$} (\#1) & $\phantom{-}2.07$ {\tiny $\pm 2.39$} (\#2) & $\phantom{-}1.23$ {\tiny $\pm 0.95$} (\#6) & $\phantom{-}0.30$ {\tiny $\pm 0.66$} (\#1) \\
SMS & $\phantom{-}0.63$ {\tiny $\pm 0.39$} (\#2) & $\phantom{-}2.16$ {\tiny $\pm 2.01$} (\#1) & $\phantom{-}1.25$ {\tiny $\pm 0.90$} (\#5) & $\phantom{-}0.16$ {\tiny $\pm 0.27$} (\#4) \\
SVS & $\phantom{-}0.52$ {\tiny $\pm 0.35$} (\#3) & $\phantom{-}1.81$ {\tiny $\pm 1.90$} (\#3) & $\phantom{-}1.15$ {\tiny $\pm 0.88$} (\#9) & $\phantom{-}0.01$ {\tiny $\pm 0.19$} (\#10) \\
Spline & $\phantom{-}0.52$ {\tiny $\pm 0.35$} (\#4) & $\phantom{-}1.62$ {\tiny $\pm 1.88$} (\#4) & $\phantom{-}1.13$ {\tiny $\pm 0.84$} (\#11) & $\phantom{-}0.21$ {\tiny $\pm 0.28$} (\#2) \\
CIR & $\phantom{-}0.51$ {\tiny $\pm 0.35$} (\#5) & $-10.34$ {\tiny $\pm 8.10$} (\#17) & $\phantom{-}1.29$ {\tiny $\pm 0.89$} (\#3) & $\phantom{-}0.17$ {\tiny $\pm 0.28$} (\#3) \\
Platt-logits & $\phantom{-}0.50$ {\tiny $\pm 0.34$} (\#6) & $\phantom{-}1.41$ {\tiny $\pm 2.17$} (\#5) & $\phantom{-}1.08$ {\tiny $\pm 0.86$} (\#14) & $\phantom{-}0.09$ {\tiny $\pm 0.18$} (\#7) \\
VS & $\phantom{-}0.50$ {\tiny $\pm 0.37$} (\#7) & $\phantom{-}0.59$ {\tiny $\pm 3.13$} (\#9) & $\phantom{-}1.14$ {\tiny $\pm 0.89$} (\#10) & $\phantom{-}0.12$ {\tiny $\pm 0.25$} (\#6) \\
Isotonic & $\phantom{-}0.45$ {\tiny $\pm 0.36$} (\#8) & $-13.85$ {\tiny $\pm 10.30$} (\#18) & $\phantom{-}1.34$ {\tiny $\pm 0.88$} (\#2) & $\phantom{-}0.07$ {\tiny $\pm 0.34$} (\#8) \\
LightGBM & $\phantom{-}0.42$ {\tiny $\pm 0.72$} (\#9) & $\phantom{-}1.07$ {\tiny $\pm 2.66$} (\#7) & $\phantom{-}1.19$ {\tiny $\pm 0.96$} (\#7) & $\phantom{-}0.13$ {\tiny $\pm 0.73$} (\#5) \\
ETS & $\phantom{-}0.42$ {\tiny $\pm 0.34$} (\#10) & $\phantom{-}0.99$ {\tiny $\pm 2.38$} (\#8) & $\phantom{-}1.16$ {\tiny $\pm 0.86$} (\#8) & $\phantom{-}0.00$ {\tiny $\pm 0.00$} (\#11) \\
TS & $\phantom{-}0.41$ {\tiny $\pm 0.33$} (\#11) & $\phantom{-}0.23$ {\tiny $\pm 3.38$} (\#10) & $\phantom{-}1.12$ {\tiny $\pm 0.84$} (\#12) & $\phantom{-}0.00$ {\tiny $\pm 0.00$} (\#11) \\
Dirichlet & $\phantom{-}0.36$ {\tiny $\pm 0.46$} (\#12) & $\phantom{-}1.20$ {\tiny $\pm 2.38$} (\#6) & $\phantom{-}1.09$ {\tiny $\pm 0.94$} (\#13) & $-0.07$ {\tiny $\pm 0.41$} (\#15) \\
Venn-Abers & $\phantom{-}0.06$ {\tiny $\pm 0.43$} (\#13) & $-0.02$ {\tiny $\pm 2.02$} (\#12) & $-0.10$ {\tiny $\pm 1.01$} (\#19) & $\phantom{-}0.02$ {\tiny $\pm 0.38$} (\#9) \\
BBQ & $\phantom{-}0.00$ {\tiny $\pm 0.43$} (\#14) & $-5.58$ {\tiny $\pm 6.61$} (\#15) & $\phantom{-}1.39$ {\tiny $\pm 0.97$} (\#1) & $-0.05$ {\tiny $\pm 0.43$} (\#14) \\
Base-model & $\phantom{-}0.00$ {\tiny $\pm 0.00$} (\#15) & $\phantom{-}0.00$ {\tiny $\pm 0.00$} (\#11) & $\phantom{-}0.00$ {\tiny $\pm 0.00$} (\#18) & $\phantom{-}0.00$ {\tiny $\pm 0.00$} (\#11) \\
XGBoost & $-0.02$ {\tiny $\pm 0.86$} (\#16) & $-1.92$ {\tiny $\pm 2.57$} (\#14) & $-0.96$ {\tiny $\pm 0.97$} (\#20) & $-0.16$ {\tiny $\pm 0.88$} (\#16) \\
MS & $-0.04$ {\tiny $\pm 0.80$} (\#17) & $-7.27$ {\tiny $\pm 15.73$} (\#16) & $\phantom{-}0.64$ {\tiny $\pm 1.07$} (\#16) & $-0.17$ {\tiny $\pm 0.53$} (\#17) \\
Hist-uniform & $-0.13$ {\tiny $\pm 0.46$} (\#18) & $-22.38$ {\tiny $\pm 16.16$} (\#20) & $\phantom{-}1.26$ {\tiny $\pm 0.93$} (\#4) & $-0.22$ {\tiny $\pm 0.42$} (\#18) \\
Kernel & $-0.32$ {\tiny $\pm 0.45$} (\#19) & $-1.12$ {\tiny $\pm 2.20$} (\#13) & $\phantom{-}0.80$ {\tiny $\pm 0.92$} (\#15) & $-0.27$ {\tiny $\pm 0.46$} (\#19) \\
Hist-quantile & $-4.39$ {\tiny $\pm 2.59$} (\#20) & $-16.86$ {\tiny $\pm 9.53$} (\#19) & $\phantom{-}0.34$ {\tiny $\pm 0.98$} (\#17) & $-4.31$ {\tiny $\pm 2.49$} (\#20) \\
\bottomrule
\end{tabular}

}
\end{table}

\begin{table}[htbp]
\centering
\caption{
Absolute post-hoc improvement for 4 metrics of interest, averaged over all experiments in the \textbf{TabArena-multiclass} post-hoc calibration benchmark.
\textbf{For readability, every value in the table is multiplied by 100.}
The $\pm$ bounds indicate standard 95\% CIs calculated using $n = $ number of experiments in the benchmark.
Methods are ranked by PHI in Brier score and we indicate ranks for each metric in parentheses.
}
\label{tab:ImprovementsTabArenaMulticlass}
\resizebox{\linewidth}{!}{
\begin{tabular}{lllll}
\toprule
Method & $\phantom{-} 100 \times \Phi_\mathrm{Brier}$ & $\phantom{-} 100 \times \Phi_\mathrm{Logloss}$ & $\phantom{-} 100 \times \Phi_\mathrm{ECE-15}$ & $\phantom{-} 100 \times \Phi_\mathrm{Accuracy}$ \\
\midrule
Spline & $\phantom{-}0.03$ {\tiny $\pm 0.05$} (\#1) & $-0.30$ {\tiny $\pm 0.21$} (\#7) & $\phantom{-}0.11$ {\tiny $\pm 0.16$} (\#6) & $\phantom{-}0.02$ {\tiny $\pm 0.06$} (\#6) \\
SMS & $\phantom{-}0.02$ {\tiny $\pm 0.04$} (\#2) & $\phantom{-}0.11$ {\tiny $\pm 0.09$} (\#2) & $\phantom{-}0.06$ {\tiny $\pm 0.14$} (\#10) & $\phantom{-}0.07$ {\tiny $\pm 0.10$} (\#4) \\
SVS & $\phantom{-}0.02$ {\tiny $\pm 0.03$} (\#3) & $\phantom{-}0.06$ {\tiny $\pm 0.08$} (\#3) & $\phantom{-}0.06$ {\tiny $\pm 0.15$} (\#11) & $\phantom{-}0.07$ {\tiny $\pm 0.09$} (\#3) \\
VS & $\phantom{-}0.00$ {\tiny $\pm 0.05$} (\#4) & $-2.59$ {\tiny $\pm 1.92$} (\#13) & $\phantom{-}0.07$ {\tiny $\pm 0.15$} (\#9) & $\phantom{-}0.10$ {\tiny $\pm 0.10$} (\#1) \\
ETS & $\phantom{-}0.00$ {\tiny $\pm 0.04$} (\#5) & $\phantom{-}0.11$ {\tiny $\pm 0.08$} (\#1) & $\phantom{-}0.18$ {\tiny $\pm 0.15$} (\#5) & $\phantom{-}0.00$ {\tiny $\pm 0.00$} (\#7) \\
Base-model & $\phantom{-}0.00$ {\tiny $\pm 0.00$} (\#6) & $\phantom{-}0.00$ {\tiny $\pm 0.00$} (\#5) & $\phantom{-}0.00$ {\tiny $\pm 0.00$} (\#14) & $\phantom{-}0.00$ {\tiny $\pm 0.00$} (\#7) \\
CIR & $-0.00$ {\tiny $\pm 0.06$} (\#7) & $-21.37$ {\tiny $\pm 6.23$} (\#18) & $\phantom{-}0.01$ {\tiny $\pm 0.17$} (\#13) & $-0.04$ {\tiny $\pm 0.09$} (\#11) \\
Platt-logits & $-0.00$ {\tiny $\pm 0.03$} (\#8) & $-0.71$ {\tiny $\pm 0.65$} (\#8) & $\phantom{-}0.10$ {\tiny $\pm 0.16$} (\#7) & $\phantom{-}0.07$ {\tiny $\pm 0.08$} (\#2) \\
Dirichlet & $-0.01$ {\tiny $\pm 0.05$} (\#9) & $-0.86$ {\tiny $\pm 0.60$} (\#11) & $-0.06$ {\tiny $\pm 0.16$} (\#15) & $\phantom{-}0.03$ {\tiny $\pm 0.12$} (\#5) \\
TS & $-0.02$ {\tiny $\pm 0.03$} (\#10) & $\phantom{-}0.05$ {\tiny $\pm 0.05$} (\#4) & $\phantom{-}0.08$ {\tiny $\pm 0.13$} (\#8) & $\phantom{-}0.00$ {\tiny $\pm 0.00$} (\#7) \\
Isotonic & $-0.06$ {\tiny $\pm 0.08$} (\#11) & $-27.65$ {\tiny $\pm 8.01$} (\#20) & $\phantom{-}0.20$ {\tiny $\pm 0.16$} (\#4) & $-0.07$ {\tiny $\pm 0.12$} (\#13) \\
MS & $-0.06$ {\tiny $\pm 0.07$} (\#12) & $-3.18$ {\tiny $\pm 2.48$} (\#15) & $\phantom{-}0.05$ {\tiny $\pm 0.18$} (\#12) & $-0.01$ {\tiny $\pm 0.13$} (\#10) \\
CatBoost & $-0.14$ {\tiny $\pm 0.11$} (\#13) & $-0.29$ {\tiny $\pm 0.18$} (\#6) & $\phantom{-}0.39$ {\tiny $\pm 0.17$} (\#1) & $-0.22$ {\tiny $\pm 0.14$} (\#16) \\
BBQ & $-0.27$ {\tiny $\pm 0.13$} (\#14) & $-5.49$ {\tiny $\pm 2.22$} (\#16) & $-0.09$ {\tiny $\pm 0.29$} (\#16) & $-0.17$ {\tiny $\pm 0.14$} (\#15) \\
Kernel & $-0.28$ {\tiny $\pm 0.18$} (\#15) & $-2.75$ {\tiny $\pm 0.85$} (\#14) & $-0.41$ {\tiny $\pm 0.34$} (\#17) & $-0.04$ {\tiny $\pm 0.10$} (\#12) \\
Venn-Abers & $-0.30$ {\tiny $\pm 0.11$} (\#16) & $-0.81$ {\tiny $\pm 0.32$} (\#10) & $-0.98$ {\tiny $\pm 0.37$} (\#19) & $-0.11$ {\tiny $\pm 0.11$} (\#14) \\
LightGBM & $-0.31$ {\tiny $\pm 0.14$} (\#17) & $-0.78$ {\tiny $\pm 0.29$} (\#9) & $\phantom{-}0.21$ {\tiny $\pm 0.20$} (\#2) & $-0.29$ {\tiny $\pm 0.18$} (\#19) \\
XGBoost & $-0.40$ {\tiny $\pm 0.13$} (\#18) & $-1.82$ {\tiny $\pm 0.24$} (\#12) & $-1.31$ {\tiny $\pm 0.26$} (\#20) & $-0.24$ {\tiny $\pm 0.17$} (\#17) \\
Hist-uniform & $-0.42$ {\tiny $\pm 0.15$} (\#19) & $-9.95$ {\tiny $\pm 3.83$} (\#17) & $\phantom{-}0.20$ {\tiny $\pm 0.18$} (\#3) & $-0.24$ {\tiny $\pm 0.12$} (\#18) \\
Hist-quantile & $-1.81$ {\tiny $\pm 0.61$} (\#20) & $-26.54$ {\tiny $\pm 7.28$} (\#19) & $-0.91$ {\tiny $\pm 0.38$} (\#18) & $-1.29$ {\tiny $\pm 0.47$} (\#20) \\
\bottomrule
\end{tabular}

}
\end{table}

\begin{table}[htbp]
\centering
\caption{
Absolute post-hoc improvement for 4 metrics of interest, averaged over all experiments in the \textbf{CV-multiclass} post-hoc calibration benchmark.
\textbf{For readability, every value in the table is multiplied by 100.}
The $\pm$ bounds indicate standard 95\% CIs calculated using $n = $ number of experiments in the benchmark.
Methods are ranked by PHI in Brier score and we indicate ranks for each metric in parentheses.
}
\label{tab:ImprovementsCVMulticlass}
\resizebox{\linewidth}{!}{
\begin{tabular}{lllll}
\toprule
Method & $\phantom{-} 100 \times \Phi_\mathrm{Brier}$ & $\phantom{-} 100 \times \Phi_\mathrm{Logloss}$ & $\phantom{-} 100 \times \Phi_\mathrm{ECE-15}$ & $\phantom{-} 100 \times \Phi_\mathrm{Accuracy}$ \\
\midrule
SMS & $\phantom{-}5.17$ {\tiny $\pm 1.82$} (\#1) & $\phantom{-}57.67$ {\tiny $\pm 24.40$} (\#1) & $\phantom{-}8.87$ {\tiny $\pm 2.90$} (\#1) & $\phantom{-}0.83$ {\tiny $\pm 0.48$} (\#1) \\
VS & $\phantom{-}4.83$ {\tiny $\pm 1.71$} (\#2) & $\phantom{-}49.19$ {\tiny $\pm 30.45$} (\#9) & $\phantom{-}8.52$ {\tiny $\pm 2.87$} (\#5) & $\phantom{-}0.62$ {\tiny $\pm 0.47$} (\#2) \\
SVS & $\phantom{-}4.80$ {\tiny $\pm 1.69$} (\#3) & $\phantom{-}56.52$ {\tiny $\pm 24.04$} (\#2) & $\phantom{-}8.84$ {\tiny $\pm 2.89$} (\#2) & $\phantom{-}0.54$ {\tiny $\pm 0.40$} (\#3) \\
Platt-logits & $\phantom{-}4.52$ {\tiny $\pm 1.66$} (\#4) & $\phantom{-}54.42$ {\tiny $\pm 24.05$} (\#3) & $\phantom{-}8.21$ {\tiny $\pm 2.67$} (\#6) & $\phantom{-}0.40$ {\tiny $\pm 0.38$} (\#4) \\
Spline & $\phantom{-}4.39$ {\tiny $\pm 1.59$} (\#5) & $\phantom{-}51.38$ {\tiny $\pm 23.54$} (\#7) & $\phantom{-}7.66$ {\tiny $\pm 2.55$} (\#7) & $\phantom{-}0.35$ {\tiny $\pm 0.23$} (\#7) \\
Isotonic & $\phantom{-}4.16$ {\tiny $\pm 1.58$} (\#6) & $-12.21$ {\tiny $\pm 61.80$} (\#17) & $\phantom{-}7.28$ {\tiny $\pm 2.65$} (\#8) & $\phantom{-}0.16$ {\tiny $\pm 0.26$} (\#9) \\
Venn-Abers & $\phantom{-}4.13$ {\tiny $\pm 1.81$} (\#7) & $\phantom{-}53.16$ {\tiny $\pm 24.52$} (\#6) & $\phantom{-}6.61$ {\tiny $\pm 3.58$} (\#12) & $\phantom{-}0.25$ {\tiny $\pm 0.26$} (\#8) \\
TS & $\phantom{-}4.09$ {\tiny $\pm 1.60$} (\#8) & $\phantom{-}53.69$ {\tiny $\pm 23.69$} (\#5) & $\phantom{-}8.58$ {\tiny $\pm 2.90$} (\#4) & $\phantom{-}0.00$ {\tiny $\pm 0.00$} (\#11) \\
CIR & $\phantom{-}4.08$ {\tiny $\pm 1.48$} (\#9) & $-11.03$ {\tiny $\pm 49.53$} (\#16) & $\phantom{-}6.34$ {\tiny $\pm 2.24$} (\#14) & $\phantom{-}0.35$ {\tiny $\pm 0.26$} (\#6) \\
ETS & $\phantom{-}4.06$ {\tiny $\pm 1.58$} (\#10) & $\phantom{-}53.87$ {\tiny $\pm 23.79$} (\#4) & $\phantom{-}8.58$ {\tiny $\pm 2.90$} (\#3) & $\phantom{-}0.00$ {\tiny $\pm 0.00$} (\#11) \\
CatBoost & $\phantom{-}3.62$ {\tiny $\pm 2.52$} (\#11) & $\phantom{-}50.34$ {\tiny $\pm 25.91$} (\#8) & $\phantom{-}6.97$ {\tiny $\pm 4.18$} (\#9) & $\phantom{-}0.37$ {\tiny $\pm 0.51$} (\#5) \\
XGBoost & $\phantom{-}3.56$ {\tiny $\pm 2.06$} (\#12) & $\phantom{-}47.96$ {\tiny $\pm 25.51$} (\#10) & $\phantom{-}6.80$ {\tiny $\pm 3.83$} (\#10) & $-0.27$ {\tiny $\pm 0.56$} (\#15) \\
LightGBM & $\phantom{-}2.64$ {\tiny $\pm 2.57$} (\#13) & $\phantom{-}41.19$ {\tiny $\pm 26.62$} (\#11) & $\phantom{-}6.51$ {\tiny $\pm 2.61$} (\#13) & $-0.51$ {\tiny $\pm 1.63$} (\#16) \\
BBQ & $\phantom{-}0.89$ {\tiny $\pm 1.10$} (\#14) & $\phantom{-}19.63$ {\tiny $\pm 31.57$} (\#14) & $\phantom{-}5.56$ {\tiny $\pm 2.40$} (\#15) & $\phantom{-}0.03$ {\tiny $\pm 0.24$} (\#10) \\
Hist-uniform & $\phantom{-}0.29$ {\tiny $\pm 1.68$} (\#15) & $-82.41$ {\tiny $\pm 91.24$} (\#19) & $\phantom{-}5.12$ {\tiny $\pm 2.61$} (\#16) & $-0.62$ {\tiny $\pm 0.73$} (\#17) \\
Kernel & $\phantom{-}0.25$ {\tiny $\pm 1.08$} (\#16) & $\phantom{-}30.67$ {\tiny $\pm 20.87$} (\#12) & $\phantom{-}3.72$ {\tiny $\pm 2.79$} (\#17) & $-0.23$ {\tiny $\pm 0.32$} (\#14) \\
Dirichlet & $\phantom{-}0.06$ {\tiny $\pm 3.84$} (\#17) & $\phantom{-}21.01$ {\tiny $\pm 35.54$} (\#13) & $\phantom{-}3.06$ {\tiny $\pm 3.54$} (\#18) & $-1.33$ {\tiny $\pm 1.67$} (\#18) \\
Base-model & $\phantom{-}0.00$ {\tiny $\pm 0.00$} (\#18) & $\phantom{-}0.00$ {\tiny $\pm 0.00$} (\#15) & $\phantom{-}0.00$ {\tiny $\pm 0.00$} (\#19) & $\phantom{-}0.00$ {\tiny $\pm 0.00$} (\#11) \\
Hist-quantile & $-5.10$ {\tiny $\pm 6.65$} (\#19) & $-16.59$ {\tiny $\pm 40.02$} (\#18) & $\phantom{-}6.69$ {\tiny $\pm 3.34$} (\#11) & $-10.92$ {\tiny $\pm 8.73$} (\#20) \\
MS & $-10.12$ {\tiny $\pm 10.41$} (\#20) & $-762.82$ {\tiny $\pm 611.50$} (\#20) & $-4.21$ {\tiny $\pm 8.06$} (\#20) & $-4.39$ {\tiny $\pm 3.62$} (\#19) \\
\bottomrule
\end{tabular}

}
\end{table}

\begin{table}[htbp]
\centering
\caption{
Absolute post-hoc improvement for 4 metrics of interest, averaged over all experiments in the \textbf{ImageNet-multiclass} post-hoc calibration benchmark.
\textbf{For readability, every value in the table is multiplied by 100.}
The $\pm$ bounds indicate standard 95\% CIs calculated using $n = $ number of experiments in the benchmark.
Methods are ranked by PHI in Brier score and we indicate ranks for each metric in parentheses.
}
\label{tab:ImprovementsImageNetMulticlass}
\resizebox{\linewidth}{!}{
\begin{tabular}{lllll}
\toprule
Method & $\phantom{-} 100 \times \Phi_\mathrm{Brier}$ & $\phantom{-} 100 \times \Phi_\mathrm{Logloss}$ & $\phantom{-} 100 \times \Phi_\mathrm{ECE-15}$ & $\phantom{-} 100 \times \Phi_\mathrm{Accuracy}$ \\
\midrule
ETS & $\phantom{-}0.42$ {\tiny $\pm 0.24$} (\#1) & $\phantom{-}2.57$ {\tiny $\pm 2.13$} (\#3) & $\phantom{-}3.84$ {\tiny $\pm 1.69$} (\#1) & $\phantom{-}0.00$ {\tiny $\pm 0.00$} (\#2) \\
SVS & $\phantom{-}0.33$ {\tiny $\pm 0.25$} (\#2) & $\phantom{-}4.10$ {\tiny $\pm 2.06$} (\#1) & $\phantom{-}3.21$ {\tiny $\pm 1.59$} (\#2) & $\phantom{-}0.00$ {\tiny $\pm 0.01$} (\#1) \\
TS & $\phantom{-}0.32$ {\tiny $\pm 0.25$} (\#3) & $\phantom{-}4.00$ {\tiny $\pm 2.06$} (\#2) & $\phantom{-}3.20$ {\tiny $\pm 1.59$} (\#3) & $\phantom{-}0.00$ {\tiny $\pm 0.00$} (\#2) \\
Platt-logits & $\phantom{-}0.11$ {\tiny $\pm 0.29$} (\#4) & $\phantom{-}0.90$ {\tiny $\pm 2.76$} (\#4) & $\phantom{-}2.89$ {\tiny $\pm 1.72$} (\#4) & $-0.23$ {\tiny $\pm 0.16$} (\#6) \\
VS & $\phantom{-}0.00$ {\tiny $\pm 0.32$} (\#5) & $-4.28$ {\tiny $\pm 3.39$} (\#6) & $\phantom{-}2.67$ {\tiny $\pm 1.55$} (\#6) & $-0.11$ {\tiny $\pm 0.11$} (\#5) \\
Base-model & $\phantom{-}0.00$ {\tiny $\pm 0.00$} (\#6) & $\phantom{-}0.00$ {\tiny $\pm 0.00$} (\#5) & $\phantom{-}0.00$ {\tiny $\pm 0.00$} (\#9) & $\phantom{-}0.00$ {\tiny $\pm 0.00$} (\#2) \\
CIR & $-0.99$ {\tiny $\pm 0.52$} (\#7) & $-354.46$ {\tiny $\pm 63.40$} (\#8) & $\phantom{-}0.01$ {\tiny $\pm 1.86$} (\#8) & $-0.51$ {\tiny $\pm 0.25$} (\#7) \\
Isotonic & $-1.46$ {\tiny $\pm 0.52$} (\#8) & $-461.61$ {\tiny $\pm 81.58$} (\#9) & $\phantom{-}0.80$ {\tiny $\pm 1.87$} (\#7) & $-1.01$ {\tiny $\pm 0.28$} (\#8) \\
Hist-uniform & $-11.04$ {\tiny $\pm 1.88$} (\#9) & $-590.18$ {\tiny $\pm 68.20$} (\#10) & $-2.90$ {\tiny $\pm 1.22$} (\#10) & $-6.02$ {\tiny $\pm 1.82$} (\#9) \\
Hist-quantile & $-65.06$ {\tiny $\pm 7.59$} (\#10) & $-305.57$ {\tiny $\pm 35.11$} (\#7) & $\phantom{-}2.84$ {\tiny $\pm 1.73$} (\#5) & $-73.73$ {\tiny $\pm 6.70$} (\#10) \\
\bottomrule
\end{tabular}

}
\end{table}

\FloatBarrier

\section{Statistical analysis}
\label{app:statistical_analysis}

\begin{figure}[htbp]
    \centering
    \begin{minipage}{\textwidth}
        \centering
        \includegraphics[width=\linewidth]{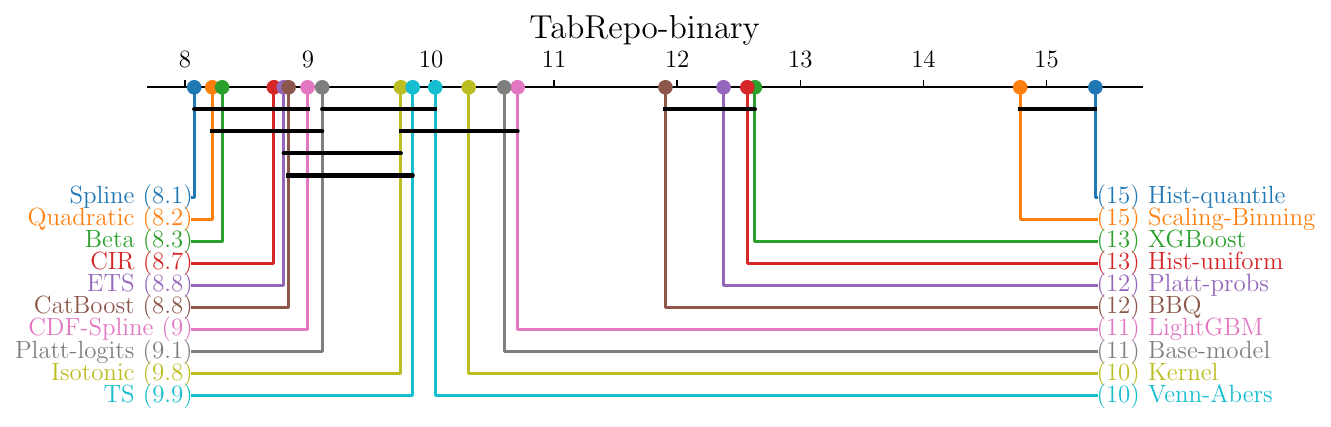}
    \end{minipage}
    \begin{minipage}{\textwidth}
        \centering
        \includegraphics[width=\linewidth]{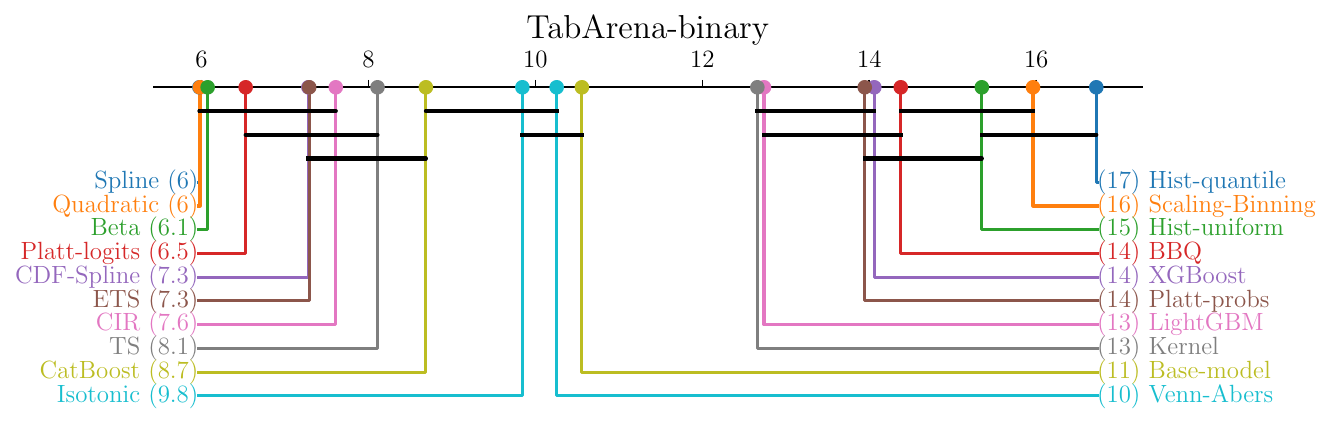}
    \end{minipage}
    \begin{minipage}{\textwidth}
        \centering
        \includegraphics[width=\linewidth]{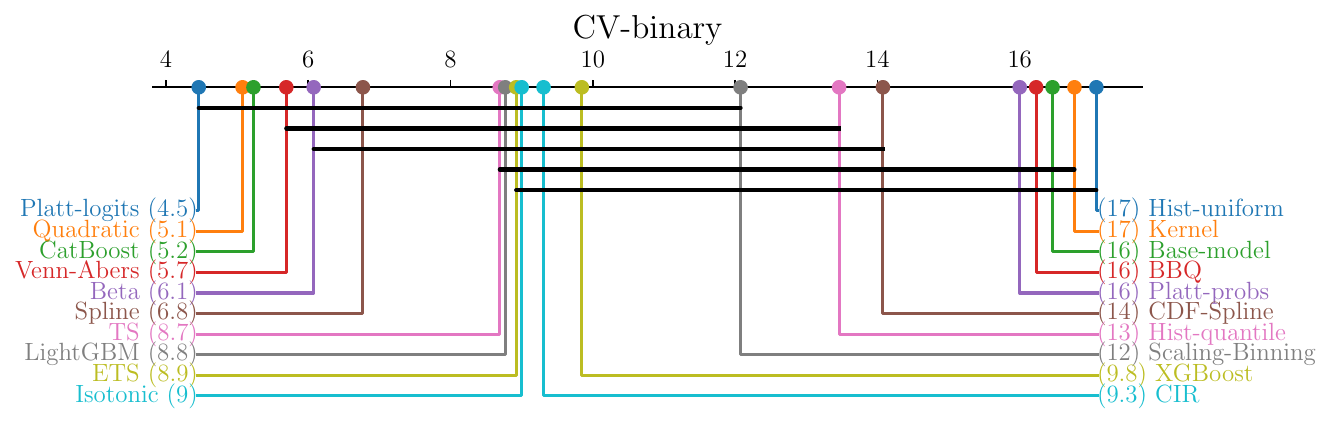}
    \end{minipage}
    \caption{
    Critical difference diagrams for \textbf{TabRepo-binary} (top), \textbf{TabArena-binary} (middle) and \textbf{CV-binary} (bottom).
    Methods are sorted by their average rank on all experiments (x-axis) and black horizontal lines connect groups of methods that are not significantly different.
    Numbers in parentheses indicate the average rank of each method (lower is better).
    }
    \label{fig:BinaryCDs}
\end{figure}

\begin{figure}[htbp]
    \centering
    \begin{minipage}{\textwidth}
        \centering
        \includegraphics[width=\linewidth]{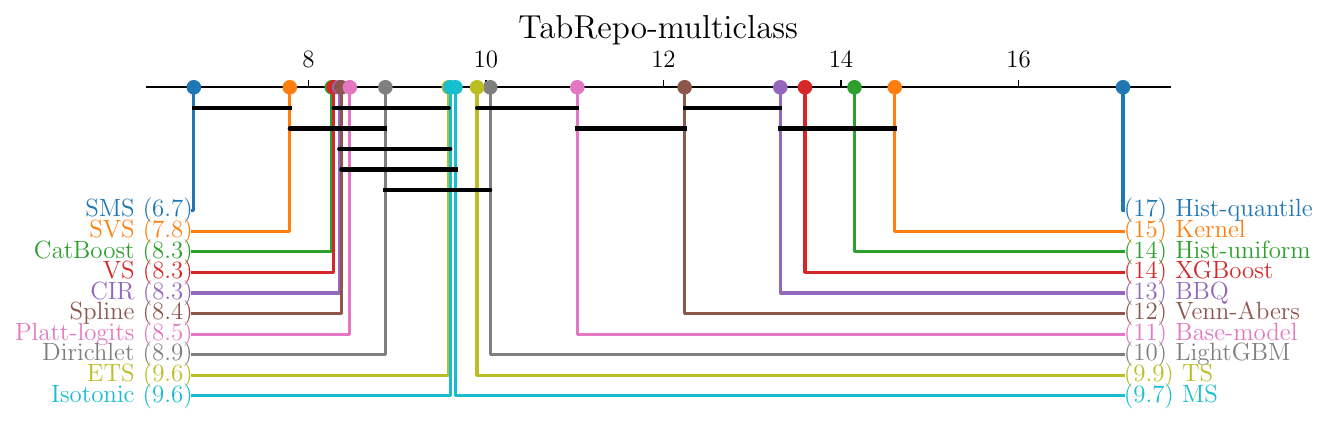}
    \end{minipage}
    \begin{minipage}{\textwidth}
        \centering
        \includegraphics[width=\linewidth]{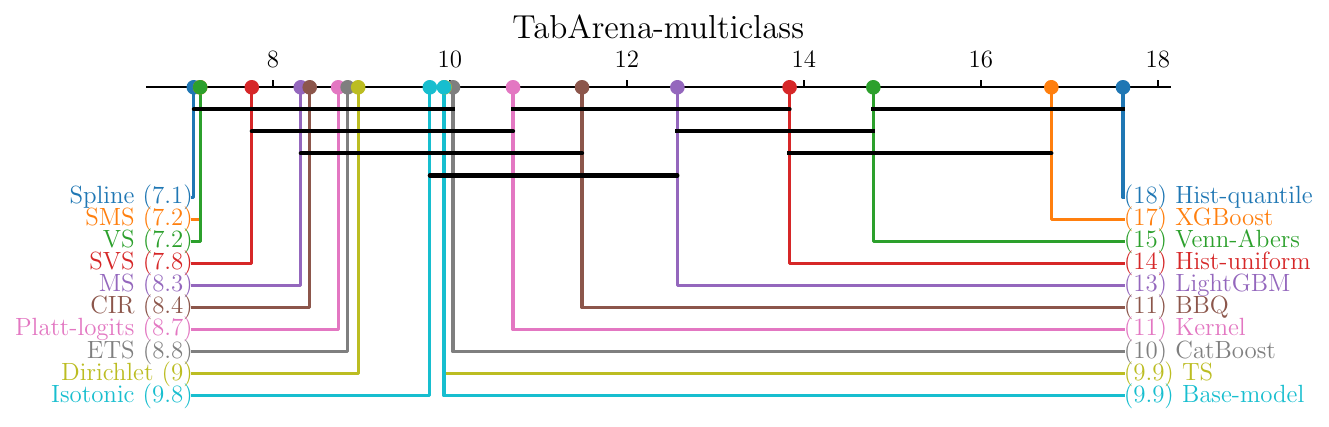}
    \end{minipage}
    \begin{minipage}{\textwidth}
        \centering
        \includegraphics[width=\linewidth]{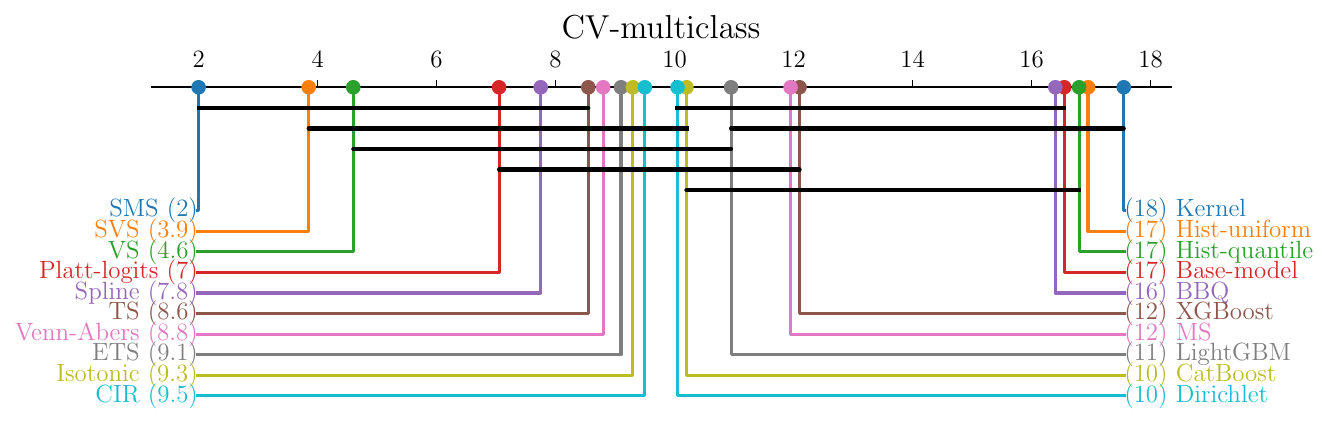}
    \end{minipage}
    \begin{minipage}{\textwidth}
        \centering
        \includegraphics[width=\linewidth]{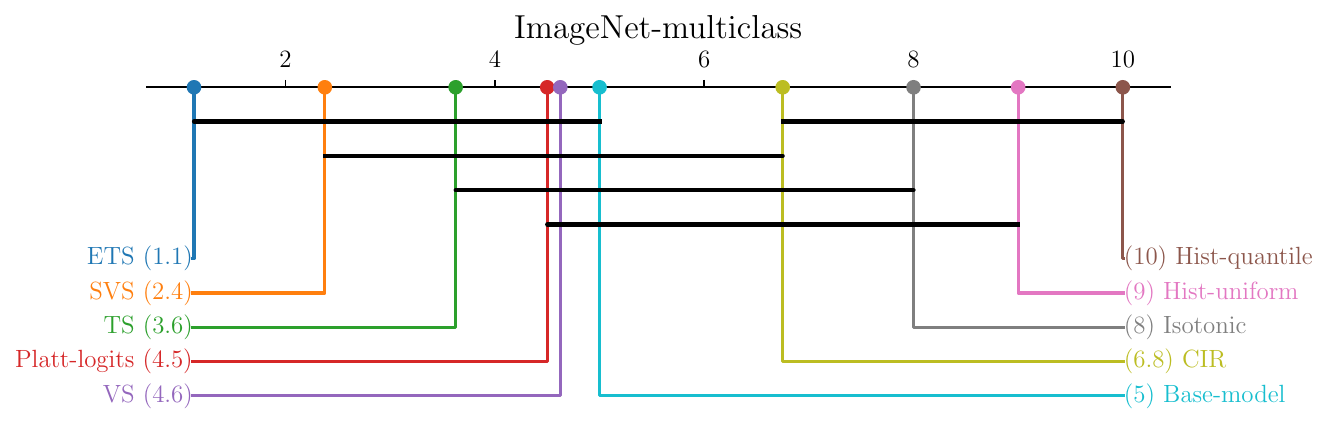}
    \end{minipage}
    \caption{
    Critical difference diagrams for \textbf{TabRepo-multiclass} (top), \textbf{TabArena-multiclass} (middle), \textbf{CV-multiclass} (bottom) and \textbf{ImageNet-multiclass} (fourth line).
    Methods are sorted by their average rank on all experiments (x-axis) and black horizontal lines connect groups of methods that are not significantly different.
    Numbers in parentheses indicate the average rank of each method (lower is better).
    }
    \label{fig:MulticlassCDs}
\end{figure}

The critical difference diagrams in \Cref{fig:BinaryCDs} and \Cref{fig:MulticlassCDs} provide statistical significance results.

\end{document}